\documentclass[letterpaper]{article} 
\usepackage{aaai2027}
\nocopyright
\usepackage[hyphens]{url}  
\usepackage{graphicx} 
\usepackage{natbib}  
\usepackage{caption} 
\usepackage{algorithm}
\usepackage{algorithmic}

\usepackage{newfloat}
\usepackage{listings}
\DeclareCaptionStyle{ruled}{labelfont=normalfont,labelsep=colon,strut=off} 
\floatstyle{ruled}
\newfloat{listing}{tb}{lst}{}
\floatname{listing}{Listing}

\usepackage{booktabs}
\usepackage{amsmath}
\usepackage{amssymb}

\usepackage{multirow}
\usepackage{makecell}
\title{Cooperative Coevolution for Resource-Constrained Agentic LLM Post-Training}
\author{
Zhiyuan Wang\textsuperscript{\rm 1}, Shengcai Liu\textsuperscript{\rm 1}, Jiahao Wu\textsuperscript{\rm 1,2}, Ning Lu\textsuperscript{\rm 1, 3}, Hui Ouyang\textsuperscript{\rm 1,5},\\ Shaofeng Zhang\textsuperscript{\rm 1,4}, Haoze Lv\textsuperscript{\rm 1,4}, Ke Tang\textsuperscript{\rm 1}\corresponding
}
\affiliations{
	\textsuperscript{\rm 1} Guangdong Provincial Key Laboratory of Brain-inspired Intelligent Computation, Department of Computer Science and Engineering, Southern University of Science and Technology\\
	\textsuperscript{\rm 2} Hong Kong Polytechnic University\quad
	\textsuperscript{\rm 3} Hong Kong University of Science and Technology\\
	\textsuperscript{\rm 4} Zhongguancun Academy\quad
	\textsuperscript{\rm 5} Shenzhen Loop Area Institute\\
	wangzy2020@mail.sustech.edu.cn, liusc3@sustech.edu.cn, \{12068041,12150066,12545007,12445025,11912814\}@mail.sustech.edu.cn, tangk3@sustech.edu.cn
}

\begin{document}

\maketitle

\begin{abstract}
Tool-using large language model~(LLM) agents produce long, multi-turn trajectories, making gradient-based post-training memory-intensive. Evolution strategies~(ES) enable memory-efficient full-parameter post-training without backpropagation and can eventually match the performance of gradient-based reinforcement learning~(RL). However, resource-constrained settings typically offer only a few GPUs, so the high GPU-hour requirements of ES translate into prohibitively long training times. To address this, we introduce Cooperative Parameter-subspace Evolution Strategy~(CoPES), a cooperative coevolutionary method that decomposes the full parameter space into lower-dimensional subspaces and searches over them cooperatively to improve optimization efficiency. On Qwen3.5-4B math, CoPES recovers 92\% of full-parameter GRPO's validation-accuracy gain within the GPU-hour budget of GRPO's best validation checkpoint, versus 67\% for standard ES. Its theoretical GPU memory requirement is less than one-eighth that of full-parameter GRPO. It consistently outperforms standard ES and LoRA-based GRPO on all evaluated pass@$k$ metrics across five math benchmarks. CoPES's gains extend to Qwen3.5-9B with unchanged optimization hyperparameters and to multi-hop QA. These results demonstrate an improved trade-off between memory requirements and training time for agentic LLM post-training under resource constraints. The code is open-sourced in \url{https://github.com/MetaronWang/CoPES}
\end{abstract}

\section{Introduction}

Tool-using large language model~(LLM) agents solve complex tasks by interleaving language generation with external actions and environmental feedback~\cite{yao2023react,JiangSQLZZY25}. Post-training with verifiable rewards offers a scalable way to improve such behavior without requiring annotated trajectories. Reinforcement learning~(RL) has consequently become a mainstream paradigm for agentic post-training, with current approaches dominated by gradient-based methods~\cite{ZhangLandscape2026}. However, unlike models solving conventional single-turn tasks, tool-using agents often produce long, multi-turn trajectories. Backpropagating through these trajectories creates a substantial GPU memory bottleneck for agentic LLM post-training.

Gradient-based post-training must store model weights, gradients, optimizer states, and the intermediate activations required for backpropagation. Prior work has proposed several approaches to reduce GPU memory usage during post-training. Group Relative Policy Optimization~(GRPO) eliminates the need for a learned critic~\cite{shao2024deepseekmath}, while low-rank adaptation~(LoRA) reduces gradient and optimizer-state memory by updating only low-rank adapters~\cite{hu2022lora}. However, neither avoids backpropagation through the policy model and the resulting need to store intermediate activations. Because activation memory grows with context length, this unresolved cost becomes especially severe for agentic post-training with long, multi-turn trajectories~\cite{korthikanti2023activation}. In practice, high-memory GPUs are costly, and obtaining them at scale is often difficult. These constraints motivate our focus on resource-constrained settings with limited per-GPU memory and only a small number of available GPUs, where the activation cost of long trajectories can make gradient-based agentic LLM post-training difficult to deploy.

Evolution strategies~(ES) alleviate this memory bottleneck by avoiding backpropagation. Standard ES evaluates randomly perturbed models using scalar rewards and aggregates the results into a parameter update, requiring only forward generation and reward evaluation~\cite{salimans2017es}. It can therefore optimize all model parameters without storing parameter gradients or backpropagation activations, substantially reducing GPU memory usage~\cite{sun2026essam}. Recent work shows that ES can eventually achieve performance comparable to gradient-based RL in LLM post-training~\cite{qiu2026esatscale}, but typically requires substantially more GPU-hours to do so~\cite{sun2026essam}. Moreover, existing studies focus primarily on single-turn reasoning tasks. In agentic settings, the high cost of generating each long, multi-turn trajectory further magnifies this computational disadvantage. Although ES evaluations are highly parallelizable, resource-constrained settings provide too few GPUs to absorb the additional GPU-hours through parallelism, resulting in impractically long wall-clock training times. The central challenge is therefore to improve the optimization efficiency of ES without sacrificing its memory advantage.

To address this challenge, we propose a cooperative coevolutionary method~\cite{potter1994cooperative} for ES-based LLM post-training, termed Cooperative Parameter-subspace Evolution Strategy~(CoPES). High dimensionality often reduces the effectiveness of evolutionary optimization, motivating cooperative coevolution to divide the search space into smaller subproblems~\cite{yang2008large}. CoPES applies this principle to full-parameter ES by cooperatively searching lower-dimensional parameter subspaces. By confining each perturbation to one subspace, each scalar reward reflects a lower-dimensional change under a shared full-model context rather than simultaneous changes across the entire parameter space, allowing the fixed evaluation budget to be used more effectively. CoPES scales perturbations by subspace dimensionality to match the expected squared norm of a full-space perturbation, jointly standardizes rewards across subspaces, and synchronously composes the resulting updates. Together, these designs improve parameter exploration under fixed evaluation and GPU-hour budgets while retaining the forward-only, memory-efficient, and full-parameter nature of standard ES.

Our contributions are threefold:
\begin{itemize}
    \item We introduce CoPES, which combines parameter-space decomposition and joint reward standardization for ES-based LLM post-training. To our knowledge, CoPES is the first cooperative coevolutionary method for full-parameter agentic LLM post-training.
    \item We show that CoPES improves fixed-budget optimization efficiency while preserving the memory-efficient, full-parameter advantages of ES, consistently outperforming standard ES and LoRA-based GRPO within the GPU-hour budget of full-parameter GRPO's best validation checkpoint.
    \item Experiments on five math and three question-answering benchmarks, together with hardware-feasibility tests and controlled ablations, validate CoPES across tasks and multiple metrics while isolating the effects of its key components.
\end{itemize}

\section{Related Work}

\subsection{Agentic LLM Post-Training}

Recent agentic reinforcement learning methods optimize multi-turn tool-use trajectories using outcome-based or verifiable rewards~\cite{ZhangLandscape2026,jiang2026verltool}. For example, Search-R1 trains LLMs to issue multiple search queries during reasoning and incorporate retrieved evidence~\cite{jin2025searchr1}, while ReTool combines a supervised cold start with outcome-driven RL to learn when and how to invoke a code interpreter~\cite{feng2025retool}. These methods commonly build on gradient-based policy optimization. GRPO removes the learned critic by estimating advantages from groups of sampled responses~\cite{shao2024deepseekmath}, and LoRA reduces the number of trainable parameters through low-rank adapters~\cite{hu2022lora}. Neither approach eliminates policy-model backpropagation. CoPES instead avoids policy-model backpropagation while optimizing all model parameters, addressing the broader GPU memory overhead of gradient-based post-training that critic-free and parameter-efficient methods only partially reduce.

\subsection{Evolution Strategies for LLM Post-Training}

Evolution strategies estimate parameter updates from the rewards of randomly perturbed models and therefore do not backpropagate through sampled trajectories~\cite{wierstra2014nes,salimans2017es}. Recent work has demonstrated that evolutionary optimization can scale directly to billion-parameter LLMs~\cite{qiu2026esatscale,liu2025ea4llm}. ESSAM combines ES with sharpness-aware maximization to improve generalization on mathematical reasoning while requiring only inference-level GPU memory~\cite{sun2026essam}. EGGROLL instead represents individual perturbations with low-rank factors to support batched evaluation at large population sizes~\cite{sarkar2026hyperscale}. Collectively, these studies establish evolutionary optimization as a viable backpropagation-free alternative for LLM post-training. Nevertheless, standard ES is computationally intensive, often requiring substantial GPU-hours~\cite{sun2026essam}. In resource-constrained settings, limited parallelism translates this computational demand into long wall-clock training times. CoPES targets this efficiency bottleneck in agentic LLM post-training through cooperative parameter-subspace searches while retaining the forward-only, full-parameter nature of standard ES.

\subsection{Cooperative Coevolution}

Cooperative coevolution was introduced to evolve complex solutions as interacting, coadapted subcomponents~\cite{potter1994cooperative,potter2000architecture}. For continuous optimization, it decomposes a high-dimensional decision vector into subspaces, optimizes the corresponding subproblems, and evaluates each partial solution within a context supplied by the remaining subcomponents. Subsequent studies developed a large-scale optimization framework with random grouping and adaptive weighting~\cite{yang2008large} and introduced differential grouping to identify interacting variables and construct subproblems that better reflect variable interdependence~\cite{omidvar2014differential}. Recent work has extended this paradigm to 1.7-million-dimensional neural policy search and scalable many-objective optimization~\cite{zhang2022ccncs,hu2024coerl,QianWQACZ25}. CoPES coordinates full-parameter updates through random repartitioning, joint reward standardization, and dimension-aware perturbation scaling.

\section{Methodology}
\label{sec:method}

\begin{figure*}[htbp]
    \centering
    \includegraphics[width=0.85\textwidth]{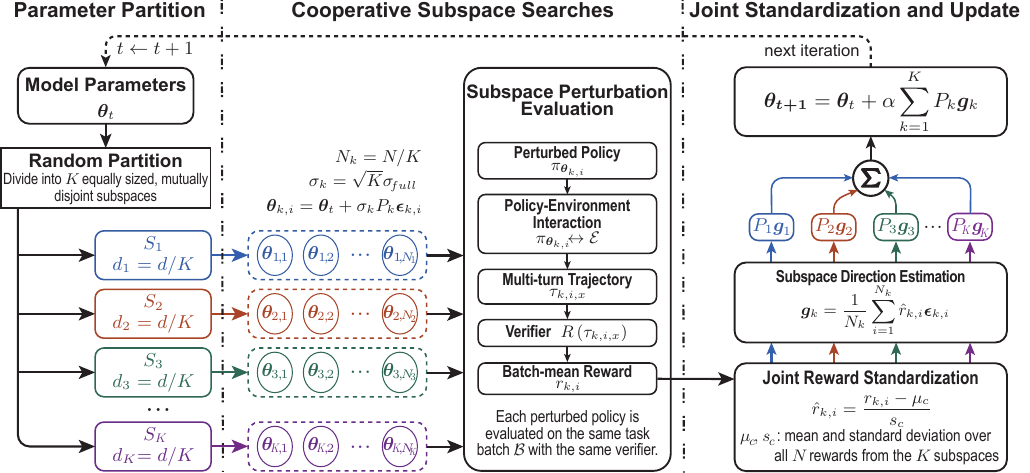}
    \caption{Overview of CoPES. At each training step, the model parameters are randomly partitioned into \(K\) disjoint subspaces, and \(N/K\) perturbations are evaluated within each subspace under a shared full-model context. Rewards from all subspaces are jointly standardized, and the resulting subspace estimates are composed into a full-parameter update.}
    \label{fig:copes_overview}
\end{figure*}

Building on cooperative coevolution~\cite{potter1994cooperative}, we propose Cooperative Parameter-subspace Evolution Strategy~(CoPES) for full-parameter ES-based agentic LLM post-training. At each training step, CoPES randomly partitions the parameter space into equally sized subspaces, allocates a fixed total of \(N\) perturbations across them, and evaluates every subspace perturbation in the context of the current full model. It then jointly standardizes the resulting rewards and composes the subspace directions into a full-parameter update. Figure~\ref{fig:copes_overview} illustrates this workflow. We first formalize agentic post-training and the standard ES update, then present the three central designs of CoPES and its memory-efficient implementation.

\subsection{Problem Formulation and Evolution Strategies}

\paragraph{Agentic Post-Training Objective.}
Given a task prompt \(x\sim\mathcal{D}\), an LLM policy \(\pi_{\boldsymbol{\theta}}\) interacts with a tool environment \(\mathcal{E}\) to produce a multi-turn trajectory \(\tau=(x,a_1,o_1,\ldots,a_T,o_T)\), where actions are sampled from the policy and observations are returned by the environment. A verifier assigns the completed trajectory a scalar reward \(R(\tau)\). The post-training objective is
\begin{equation}
J(\boldsymbol{\theta})=\mathbb{E}_{x\sim\mathcal{D},\tau\sim(\pi_{\boldsymbol{\theta}},\mathcal{E})(\cdot\mid x)}\left[R(\tau)\right],
\label{eq:agentic_objective}
\end{equation}
where \(\boldsymbol{\theta}\in\mathbb{R}^{d}\) contains all trainable model parameters. The post-training algorithm only observes completed trajectories and their scalar rewards; it does not require gradients through either the environment or the reward function. For an optimization algorithm \(\mathcal{A}\) and a GPU-hour budget \(B\), let
\begin{equation}
\boldsymbol{\theta}_{B}=\mathcal{A}(\boldsymbol{\theta}_{0},\mathcal{D},B)
\end{equation}
denote the parameters obtained within that budget. Our objective is to maximize \(J(\boldsymbol{\theta}_{B})\) under a fixed GPU-hour budget \(B\). The experimental protocol defines how \(B\) is measured across methods, while the derivations below use the per-step perturbation budget to compare standard ES and CoPES.

\paragraph{Standard Evolution Strategy.}
Following the one-sided ES update adopted for full-parameter LLM post-training~\cite{salimans2017es,qiu2026esatscale}, standard ES samples \(N\) independent Gaussian directions for the current parameters \(\boldsymbol{\theta}\) and constructs
\begin{equation}
\boldsymbol{\epsilon}_{i}\sim\mathcal{N}(\boldsymbol{0},\boldsymbol{I}_{d}),
\qquad
\boldsymbol{\theta}_{i}=\boldsymbol{\theta}+\sigma\boldsymbol{\epsilon}_{i},
\end{equation}
where \(\sigma\) is the full-space perturbation scale. Each perturbed model generates one trajectory \(\tau_{i,x}\) for every \(x\in\mathcal{B}\), producing the batch-mean reward
\begin{equation}
r_i=\frac{1}{|\mathcal{B}|}\sum_{x\in\mathcal{B}} R(\tau_{i,x}).
\label{eq:batch_mean_reward}
\end{equation}
ES standardizes these \(N\) rewards using
\begin{equation}
\begin{gathered}
\mu_r=\frac{1}{N}\sum_{i=1}^{N}r_i,
\qquad
s_r=\sqrt{\frac{1}{N}\sum_{i=1}^{N}(r_i-\mu_r)^2},\\
\hat r_i=\frac{r_i-\mu_r}{s_r},
\end{gathered}
\end{equation}
and applies
\begin{equation}
\boldsymbol{\theta}^{+}=\boldsymbol{\theta}+\frac{\alpha}{N}\sum_{i=1}^{N}\hat r_i\boldsymbol{\epsilon}_{i},
\label{eq:standard_es_update}
\end{equation}
where \(\alpha\) is the effective update step size and is set independently of \(\sigma\). This update uses no mirrored samples, unperturbed-model baseline, momentum, or gradient-based optimizer. With a limited population, each sampled direction spans the entire \(d\)-dimensional parameter space and is evaluated through a single scalar reward. This motivates decomposing the search while keeping the total number of perturbation evaluations fixed.

\subsection{Cooperative Parameter-Subspace Evolution Strategy}
\paragraph{Cooperative Subspace Search.}
CoPES replaces the monolithic \(d\)-dimensional search with \(K\) cooperative searches over lower-dimensional subspaces. At every training step, it samples a uniform random partition \(\{\mathcal{S}_{k}\}_{k=1}^{K}\) of all parameter indices such that
\begin{equation}
\begin{gathered}
\mathcal{S}_{k}\cap\mathcal{S}_{\ell}=\varnothing\ (k\neq\ell),
\quad
\bigcup_{k=1}^{K}\mathcal{S}_{k}=\{1,\ldots,d\},\\
d_k=|\mathcal{S}_k|=\frac{d}{K}. 
\end{gathered}
\end{equation}
We assume that \(d\) and \(N\) are divisible by \(K\) to simplify notation. The partition is resampled each step so that parameters are not permanently restricted to a fixed grouping. Let \(P_k:\mathbb{R}^{d_k}\rightarrow\mathbb{R}^{d}\) embed a subspace vector into the full parameter space. CoPES allocates \(N_k=N/K\) perturbations to each subspace and constructs
\begin{equation}
\boldsymbol{\epsilon}_{k,i}\sim\mathcal{N}(\boldsymbol{0},\boldsymbol{I}_{d_k}),
\qquad
\boldsymbol{\theta}_{k,i}=\boldsymbol{\theta}+\sigma_k P_k\boldsymbol{\epsilon}_{k,i}.
\label{eq:subspace_perturbation}
\end{equation}
Only parameters in \(\mathcal{S}_k\) are perturbed; all remaining parameters retain their current values and provide a shared full-model context. Each perturbed model is evaluated on the same batch \(\mathcal{B}\), yielding
\begin{equation}
r_{k,i}=\frac{1}{|\mathcal{B}|}\sum_{x\in\mathcal{B}}R(\tau_{k,i,x}).
\label{eq:subspace_reward}
\end{equation}
Although each subspace receives only \(N/K\) perturbations, its search dimension is reduced from \(d\) to \(d/K\). More importantly, each scalar reward is associated with a perturbation confined to one subspace under a shared full-model context, rather than with simultaneous perturbations across the entire parameter space. Because the subspaces have disjoint supports, CoPES obtains \(K\) mutually orthogonal subspace estimates and composes them into a full-parameter update.

\paragraph{Dimension-Aware Perturbation Scale.}
Using the full-space scale unchanged would reduce the magnitude of a subspace perturbation simply because it contains fewer dimensions. For \(\boldsymbol{\delta}=\sigma\boldsymbol{\epsilon}\) and \(\boldsymbol{\delta}_{k}=\sigma_kP_k\boldsymbol{\epsilon}_{k}\), their expected squared norms are
\begin{equation}
\begin{aligned}
\mathbb{E}\!\left[\|\boldsymbol{\delta}\|_2^2\right]
=d\sigma^2,\quad
\mathbb{E}\!\left[\|\boldsymbol{\delta}_{k}\|_2^2\right]
=d_k\sigma_k^2.
\end{aligned}
\end{equation}
Matching these quantities gives
\begin{equation}
\sigma_k=\sigma\sqrt{\frac{d}{d_k}}=\sqrt{K}\,\sigma.
\label{eq:dimension_aware_scale}
\end{equation}
This adjustment matches the expected squared perturbation norm of full-space ES while retaining the same update step size \(\alpha\).

\paragraph{Joint Reward Standardization and Cooperative Update.}
After evaluating all \(N=\sum_k N_k\) perturbed models, CoPES pools their batch-mean rewards and computes
\begin{equation}
\begin{gathered}
\mu_c=\frac{1}{N}\sum_{k=1}^{K}\sum_{i=1}^{N_k}r_{k,i},
s_c=\sqrt{\frac{1}{N}\sum_{k=1}^{K}\sum_{i=1}^{N_k}(r_{k,i}-\mu_c)^2},\\
\hat{r}_{k,i}=\frac{r_{k,i}-\mu_c}{s_c}.
\end{gathered}
\label{eq:joint_standardization}
\end{equation}
The shared training batch, matched expected squared perturbation norm, and common reward function provide a consistent evaluation basis across subspaces. Joint standardization uses all \(N\) rewards to place centered block updates on a common scale, preserving their relative contributions. Independent standardization rescales each block separately. ES and CoPES skip updates when reward standard deviation is $\leq10^{-8}$.

CoPES estimates each subspace direction and composes the full update as
\begin{equation}
\begin{aligned}
\boldsymbol{g}_{k}=\frac{1}{N_k}\sum_{i=1}^{N_k}\hat r_{k,i}\boldsymbol{\epsilon}_{k,i},
\quad
\boldsymbol{\theta}^{+}=\boldsymbol{\theta}+\alpha\sum_{k=1}^{K}P_k\boldsymbol{g}_{k}.
\end{aligned}
\label{eq:copes_update}
\end{equation}
All perturbed models are evaluated and their rewards are standardized before any parameter update is applied. Consequently, every \(\boldsymbol{g}_k\) is estimated from the same pre-update model, and their composition is a synchronous full-parameter update at the algorithmic level. Algorithm~\ref{alg:copes} summarizes one CoPES training step. Supplementary~H provides a theoretical analysis of CoPES.

\begin{algorithm}[htbp]
\caption{CoPES}
\label{alg:copes}
\textbf{Input}: Parameters \(\boldsymbol{\theta}\), shared batch \(\mathcal{B}\), reward \(R\), population \(N\), and subspace count \(K\)\\
\textbf{Parameters}: Full-space scale \(\sigma\) and step size \(\alpha\)\\
\textbf{Output}: Updated parameters \(\boldsymbol{\theta}^{+}\)
\begin{algorithmic}[1]
\STATE Sample an equal-size random partition \(\{\mathcal{S}_k\}_{k=1}^{K}\)
\STATE Set \(d_k=d/K\), \(N_k=N/K\), and \(\sigma_k=\sqrt{K}\sigma\)
\FOR{\(k=1,\ldots,K\)}
    \FOR{\(i=1,\ldots,N_k\)}
        \STATE Sample \(\boldsymbol{\epsilon}_{k,i}\sim\mathcal{N}(\boldsymbol{0},\boldsymbol{I}_{d_k})\)
        \STATE Set \(\boldsymbol{\theta}_{k,i}=\boldsymbol{\theta}+\sigma_kP_k\boldsymbol{\epsilon}_{k,i}\)
        \STATE Generate \(\tau_{k,i,x}\) for every \(x\in\mathcal{B}\)
        \STATE Compute \(r_{k,i}\) using Equation~\eqref{eq:subspace_reward}
    \ENDFOR
\ENDFOR
\STATE Jointly standardize all \(r_{k,i}\) using Equation~\eqref{eq:joint_standardization}
\FOR{\(k=1,\ldots,K\)}
    \STATE Compute \(\boldsymbol{g}_k\) using Equation~\eqref{eq:copes_update}
\ENDFOR
\STATE \(\boldsymbol{\theta}^{+}\leftarrow\boldsymbol{\theta}+\alpha\sum_{k=1}^{K}P_k\boldsymbol{g}_k\)
\STATE \textbf{return} \(\boldsymbol{\theta}^{+}\)
\end{algorithmic}
\end{algorithm}

\subsection{Memory-Efficient Implementation and Cost}

\paragraph{Seed Replay, Chunked Processing, and Weight Backup.}
Following OpenAI-ES~\cite{salimans2017es}, CoPES represents each perturbation by a random seed and regenerates its Gaussian direction during the update, avoiding storage of full perturbation vectors; model parameters are processed in chunks. CoPES backs up pre-perturbation weights in CPU memory and restores them after each evaluation. This avoids rounding errors from subtracting replayed perturbations and keeps every candidate anchored to the same pre-update model. CoPES also stores a partition seed and replays the chunk-to-subspace assignments, avoiding a parameter-level mask. These mechanisms support stable, memory-efficient execution at LLM scale.

\paragraph{Evaluation and Memory Costs.}
CoPES matches standard ES in using \(N\) perturbed-model evaluations and the same number of generated trajectories per step; each perturbed model generates a trajectory for every prompt in \(\mathcal{B}\). Thus, decomposition does not reduce the cost of an individual evaluation but reallocates the fixed perturbation budget across lower-dimensional subspaces. Like standard ES, CoPES requires only forward generation and reward evaluation, storing no parameter gradients, optimizer states, or backpropagation activations.

\section{Experiments}
\label{sec:experiments}

We organize the main experiments around three research questions. \textbf{RQ1:} In agentic LLM post-training, can ES-based methods match gradient-based RL in performance, and how do their memory efficiency and GPU-hour cost compare? \textbf{RQ2:} Under resource constraints, how does CoPES compare with standard ES and GRPO variants in hardware feasibility and fixed-budget performance? \textbf{RQ3:} Which design choices account for CoPES's gains under a fixed evaluation budget? We study these questions on five math benchmarks, then examine whether CoPES's gains extend to a larger model and multi-hop QA.

\subsection{Experimental Setup}

\paragraph{Models and Tasks.}
We post-train Qwen3.5-4B~\cite{qwen35} on the math task and multi-hop question-answering~(QA) task. Our main experiments focus on math, while the QA experiments examine the applicability of CoPES to a different task. For math, we post-train on the MATH training set~\cite{hendrycks2021math} and evaluate on AIME 2024\footnote{\url{https://huggingface.co/datasets/HuggingFaceH4/aime_2024}}, AIME 2025\footnote{\url{https://huggingface.co/datasets/test-time-compute/aime_2025}}, GSM8K~\cite{cobbe2021gsm8k}, MATH-500~\cite{lightman2024verify}, and MATH-Test; MATH-500 is a curated subset of MATH-Test. For QA, we post-train on the HotpotQA training set and evaluate on HotpotQA, 2Wiki, and MuSiQue~\cite{yang2018hotpotqa,ho2020twowiki,trivedi2022musique}. Dataset descriptions and split details are provided in Supplementary~A.

\paragraph{Baselines and Hyperparameters.}
We compare CoPES with Qwen3.5-4B and Qwen3.5-9B without post-training, full-parameter GRPO~\cite{shao2024deepseekmath}, LoRA-based GRPO~\cite{hu2022lora}, and standard ES. Following recommended settings from prior work, all post-training methods use a prompt batch size of \(64\). Full-parameter and LoRA-based GRPO use learning rates of \(1\times10^{-6}\) and \(5\times10^{-6}\), respectively~\cite{shao2024deepseekmath,devulapalli2025clinicalgrpo}. Standard ES and CoPES both use \(N=40\) perturbations per step, \(\sigma=1\times10^{-3}\), and \(\alpha=5\times10^{-4}\)~\cite{qiu2026esatscale,liu2025ea4llm}. CoPES additionally uses \(K=4\), yielding \(\sigma_k=\sqrt{K}\sigma=2\times10^{-3}\). We evaluate model performance on a held-out validation set every 8 steps and select each method's best observed validation checkpoint. Full-parameter GRPO's selected step defines the fixed budget in RQ2. Other hyperparameters and implementation details are provided in Supplementary~A.

\paragraph{Rewards and Metrics.}
The base task score is answer accuracy for math and F1 for QA. For both tasks, the training reward additionally includes a tool-use bonus for a positive task score and format penalties. We report pass@\(k\)~\cite{chen2021codex} for math based on repeated sampling. For QA, we report accuracy, exact match~(EM), and F1, each averaged over 32 samples per problem. Since our 4B experiments were conducted on servers equipped with different GPU types, we report class-level GPU-hour estimates. We profile CoPES and full-parameter GRPO for 32 steps on the same machine as representatives of the ES-based and RL-based method classes, respectively, and extrapolate their average per-step costs to the reported training steps. Complete reward definitions, metric calculations, sampling protocols, and GPU-hour accounting are provided in Supplementary~A. Supplementary~C provides a detailed training-time analysis.

\paragraph{Agent and Compute Environment.}
We use a tool-using agent equipped with Local Wiki Search over Wiki18~\cite{jin2025searchr1} and a Python Sandbox. GPU-hour profiling is performed on a server with \(8\times\) RTX 5880 Ada GPUs. To accelerate training and evaluation, we additionally use servers equipped with NVIDIA A30, RTX A6000, and A100 GPUs.

\subsection{Memory and Computational Efficiency of ES}
\begin{table*}[htbp]
\centering
\setlength{\tabcolsep}{1mm}
\begin{tabular}{rccccccccc}
    \toprule
     & \multicolumn{5}{c}{Pass@1 (\%) on Benchmarks} & \multirow{2}[2]{*}{\shortstack{Train\\Steps}} & \multirow{2}[2]{*}{\shortstack{Est. GPU-\\hours}} & \multirow{2}[2]{*}{\shortstack{Valid\\Acc.}} & \multirow{2}[2]{*}{\shortstack{Run on a\\24GB GPU}} \\
\cmidrule{2-6}          & AIME 2024 & AIME 2025 & GSM8K & MATH-500 & MATH-Test &       &       &  \\
    \midrule
    \multicolumn{9}{l}{\textit{Reference Model}} \\
    {Qwen3.5-4B} & 17.14 & 35.04 & 9.22  & 57.44 & 59.02 & - & -  & 49.20   & - \\
    \midrule
    \multicolumn{9}{l}{\textit{Validation-Selected Checkpoints}} \\
    {\(+\) Full-param GRPO} & 44.28 & 58.68 & 94.19 & 89.71 & \textbf{91.66} & 16 & 78.61 & 91.47  & \(\times\) \\
    {\(+\) LoRA-based GRPO} & 41.86 & 54.75 & \textbf{94.38} & 89.34 & 91.40 & 64  & 314.45 & 91.07 & \(\times\) \\
    {\(+\) Standard ES} & \textbf{45.26} & \textbf{61.69} & 94.02 & \textbf{89.80} & 91.47 & 112 & 480.60 & \textbf{91.73}  & \(\surd\) \\
    \midrule
    \multicolumn{9}{l}{\textit{Post-Training for 16 Update Steps (Comparable Estimated GPU-hours)}} \\
    {\(+\) LoRA-based GRPO} & 28.14 & 40.76 & 47.49 & 77.19 & 79.18 & 16 & 78.61 & 73.60 & \(\times\) \\
    {\(+\) Standard ES} & 31.33 & 44.45 & 46.88 & 78.88 & 80.43 & 16  & 68.65 & 77.60 & \(\surd\) \\
    {\(+\) CoPES (ours)} & \textbf{38.39} & \textbf{52.75} & \textbf{92.57} & \textbf{88.29} & \textbf{90.16} & 16 & 68.65 & \textbf{88.00}  & \(\surd\) \\
    \midrule
    \multicolumn{9}{l}{\textit{Larger-model Reference (No Additional Post-Training)}} \\
    {Qwen3.5-9B} & 16.38 & 31.98 & 51.23 & 77.73 & 81.24 & - & -   & -     & - \\
    \bottomrule
\end{tabular}
\caption{Pass@1 (\%) and training costs on five math benchmarks. The 16-step block matches update count and unique prompt batches; estimated GPU-hours are 78.61 for the GRPO variants and 68.65 for the ES variants. It is therefore not an exact GPU-hour-matched comparison. MATH-500 is a subset of MATH-Test.}
\label{table: math_pass1}
\end{table*}

\paragraph{Effectiveness in Agentic Post-Training.}
Table~\ref{table: math_pass1} shows that standard ES performs comparably to the RL-based methods at its selected checkpoint, demonstrating its viability for agentic post-training. Standard ES obtains the highest pass@1 on three of the five math benchmarks.

\paragraph{Theoretical Memory Efficiency.}
On Qwen3.5-4B at 128K context length, even LoRA-based GRPO theoretically needs over \(30\times\) ES\textquotesingle s GPU memory. Supplementary~B provides the theoretical accounting and context-length curves.

\begin{figure*}[t]
    \centering
    \includegraphics[width=0.85\textwidth]{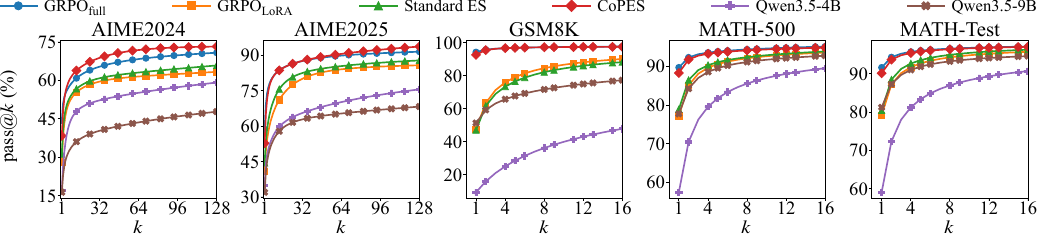}
    \caption{Pass@\(k\) of four post-training methods under the 16-step full-parameter GRPO budget, with Qwen3.5-4B and Qwen3.5-9B without post-training as references.}
    \label{fig:math_passk}
\end{figure*}

\paragraph{Computational Efficiency.}
Despite competitive performance and memory efficiency, standard ES reaches its selected checkpoint at substantially higher computational cost. As shown in Table~\ref{table: math_pass1}, standard ES is estimated to require 480.60 GPU-hours, more than six times the estimated 78.61 GPU-hours for full-parameter GRPO. In resource-constrained settings with only one or two GPUs, this computational demand translates into impractically long wall-clock training times.

Answering RQ1, standard ES can match RL-based performance with substantially less GPU memory but at considerably greater GPU-hour cost.

\begin{table}[t]
\centering
\setlength{\tabcolsep}{1mm}
\begin{tabular}{lcccc}
\toprule
Method & \shortstack{1\(\times\)\\24GB} & \shortstack{1\(\times\)\\48GB} & \shortstack{4\(\times\)\\48GB} & \shortstack{8\(\times\)\\48GB} \\
\midrule
Full-param GRPO & \(\times\) & \(\times\) & \(\times\) & \(\surd\) \\
LoRA-based GRPO & \(\times\) & \(\times\) & \(\times\) & \(\surd\)\\
Standard ES / CoPES & \(\surd\) & \(\surd\) & \(\surd\) & \(\surd\) \\
\bottomrule
\end{tabular}
\caption{Training feasibility with a 128K context limit. \(\surd\) denotes successful training, whereas \(\times\) indicates an out-of-memory failure.
}
\label{tab:hardware_feasibility}
\end{table}

\subsection{Feasibility and Performance under Resource Constraints}

\paragraph{Feasibility across Hardware Configurations.}
We evaluate training feasibility on actual hardware with all memory optimizations supported by the respective frameworks enabled; implementation details are provided in Supplementary~A. As shown in Table~\ref{tab:hardware_feasibility}, the ES-based methods can be trained on a single 24 GB GPU, whereas, among the evaluated configurations under the 128K context setting, both GRPO variants succeed only on \(8\times48\) GB GPUs. The smallest tested configuration supporting ES therefore provides less than one-eighth of the aggregate GPU memory capacity of the smallest tested configuration on which GRPO succeeds. Thus, when only one or a few GPUs are available, the GRPO-based methods are infeasible while the ES-based methods remain trainable.

\paragraph{Matched-step performance.}
In resource-constrained settings, a limited number of GPUs also limits the GPU-hours that can be spent within an acceptable wall-clock time. Full-parameter GRPO reaches its best observed validation checkpoint at step 16, as shown in Table~\ref{table: math_pass1}. We therefore train all four post-training methods for 16 steps to ensure that they process the same number of training prompts while keeping their GPU-hour costs comparable. The ES-based methods have slightly lower estimated GPU-hour costs than the RL-based methods, but the difference is small. Under this budget, Table~\ref{table: math_pass1} and Figure~\ref{fig:math_passk} show that CoPES consistently outperforms standard ES and LoRA-based GRPO across the five benchmarks while remaining competitive with full-parameter GRPO. Relative to Qwen3.5-4B without post-training, CoPES recovers 92\% of the validation-accuracy gain achieved by full-parameter GRPO, compared with 67\% for standard ES. The advantage over ES and LoRA persists across three runs (Supplementary~G).

\paragraph{Larger-\(k\) Results on AIME.}
Figure~\ref{fig:math_passk} shows that, as \(k\) increases, CoPES surpasses full-parameter GRPO on the challenging AIME 2024 and AIME 2025 benchmarks despite using the same fixed 16-step training budget.

Answering RQ2, CoPES retains the single-GPU feasibility of ES while outperforming standard ES and LoRA-based GRPO under the fixed budget; at larger \(k\), it also surpasses full-parameter GRPO on AIME.

\subsection{Ablation Studies}

\paragraph{The Number of Subspaces.}
Holding all other settings fixed, we compare \(K\in\{1,2,4,8\}\) using pass@1 on the five benchmarks after 16 training steps. As \(K\) changes, we set \(N_k=N/K\) and adjust the perturbation scale as \(\sigma_k=\sqrt{K}\sigma\). Table~\ref{tab:ablation_results} shows that performance improves as \(K\) increases from 1 to 4, with no further overall improvement at \(K=8\): \(K=4\) leads on four benchmarks, while \(K=8\) leads on GSM8K.

\paragraph{Joint Reward Standardization.}
The Indep. Z-score row in Table~\ref{tab:ablation_results} reports the results when perturbation rewards are standardized independently within each subspace. Joint reward standardization achieves higher pass@1 on every benchmark, validating its effectiveness.

\paragraph{Perturbation-Scale and Learning-Rate Controls.}
To distinguish cooperative search from changes in perturbation scale or learning rate, we compare CoPES with ES-\(2\sigma\) and ES-\(2\alpha\) under the same 16-step budget. The ES-\(2\sigma\) row in Table~\ref{tab:ablation_results} shows that increasing the perturbation scale of standard ES degrades its performance across the benchmarks rather than improving it. CoPES also outperforms ES with a doubled learning rate ($\alpha=10^{-3}$, ES-$2\alpha$) on all five benchmarks under the same 16-step budget.

\begin{table}[t]
\centering
{
\setlength{\tabcolsep}{1.0mm}
\begin{tabular}{lccccc}
\toprule
Variant & \shortstack{AIME\\2024} & \shortstack{AIME\\2025} & \shortstack{GSM\\8K} & \shortstack{MATH\\500} & \shortstack{MATH\\Test} \\
\midrule
\shortstack[l]{\(K=1\) (Std. ES)} & 31.33 & 44.45 & 46.88 & 78.88 & 80.43 \\
\(K=2\)                         & 34.48 & 46.98 & 82.07 & 85.96 & 87.70 \\
\shortstack[l]{\(K=4\) (Default)}   & \textbf{38.39} & \textbf{52.75} & 92.57 & \textbf{88.29} & \textbf{90.16} \\
\(K=8\)                         & 35.83 & 50.16 & \textbf{93.46} & 86.48 & 88.82 \\
Indep. Z-score                     & 36.45 & 50.78 & 90.98 & 87.68 & 89.45 \\
ES-\(2\sigma\)                  & 24.56 & 41.28 & 15.12 & 65.28 & 66.96 \\
ES-\(2\alpha\) & 33.31 & 48.35 & 88.10 & 85.64 & 88.19 \\
\bottomrule
\end{tabular}
}
\caption{Ablation results after 16 training steps (pass@1, \%). \(K\) denotes the number of subspaces; Indep. Z-score uses per-subspace reward standardization; ES-\(2\sigma\) is standard ES with twice the perturbation scale; ES-\(2\alpha\) doubles the learning rate.}
\label{tab:ablation_results}
\end{table}

Answering RQ3, a moderate subspace count and joint reward standardization improve CoPES, while perturbation-scale and learning-rate controls show that neither adjustment alone reproduces its gains.

\subsection{Additional Experiments on Qwen3.5-9B}

For Qwen3.5-9B, all four methods use a common 24-step budget set by full-parameter GRPO's best validation checkpoint. With unchanged 4B optimization hyperparameters, CoPES outperforms ES and LoRA-GRPO in pass@1 on all five math benchmarks (Table~\ref{tab:math_9b}), with AIME24/25 gains of 7.84/10.48 percentage points over ES. Full pass@$k$ curves appear in Supplementary~F.

\begin{table}[htbp]
\centering\small
\setlength{\tabcolsep}{2pt}
\begin{tabular}{lccccc}\toprule
Method & \shortstack{AIME\\2024} & \shortstack{AIME\\2025} & GSM8K & \shortstack{MATH\\500} & \shortstack{MATH\\Test}\\\midrule
Qwen3.5-9B & 16.38 & 31.98 & 51.23 & 77.73 & 81.24 \\
GRPO & \textbf{59.93} & \textbf{75.73} & \textbf{95.76} & \textbf{92.63} & \textbf{94.91} \\
LoRA & 41.37 & 53.28 & 95.39 & 88.74 & 91.11 \\
Std. ES & 42.15 & 55.14 & 94.50 & 88.68 & 91.02 \\
CoPES & \underline{49.99} & \underline{65.63} & \underline{95.55} & \underline{90.74} & \underline{92.84} \\
\bottomrule\end{tabular}
\caption{Math pass@1 (\%) before and after Qwen3.5-9B post-training under a 24-step budget. LoRA denotes LoRA-based GRPO. Bold and underlining mark the best and second-best results.}
\label{tab:math_9b}
\end{table}

\subsection{Additional Experiments on Multi-hop QA}

With Qwen3.5-4B, we further examine whether CoPES's effectiveness extends beyond math. Full-parameter GRPO's best observed validation checkpoint is at step 48, defining the common 48-step post-training budget. All other method-specific hyperparameters remain the same as in the math task. As shown in Table~\ref{tab:qa_results}, all three post-training methods improve accuracy and substantially increase EM and F1 over Qwen3.5-4B across the multi-hop QA benchmarks.

\begin{table}[t]
\centering
\setlength{\tabcolsep}{1mm}
\begin{tabular}{llcccc}
\toprule
Benchmark & Metric & Qwen & GRPO & Std. ES & CoPES \\
\midrule
\multirow{3}{*}{2Wiki}
  & Acc. & 44.31 & \textbf{60.95} & 55.59 & \underline{59.06} \\
  & EM    & 12.23 & \textbf{52.16} & 45.28 & \underline{48.29} \\
  & F1    & 17.12 & \textbf{58.49} & 51.53 & \underline{55.50} \\
\midrule
\multirow{3}{*}{HotpotQA}
  & Acc. & 51.75 & \textbf{53.87} & 53.22 & \underline{53.30} \\
  & EM    & 14.40 & \textbf{47.33} & 46.52 & \underline{46.96} \\
  & F1    & 22.18 & \textbf{59.38} & 58.22 & \underline{58.83} \\
\midrule
\multirow{3}{*}{MuSiQue}
  & Acc. & 25.07 & \textbf{29.89} & 29.07 & \underline{29.68} \\
  & EM    & 6.89 & \underline{23.07} & 21.55 & \textbf{23.34} \\
  & F1    & 12.08 & \underline{33.28} & 31.70 & \textbf{33.73} \\
\bottomrule
\end{tabular}
\caption{Multi-hop QA performance (\%). Qwen: Qwen3.5-4B without post-training; GRPO: full-parameter GRPO; Std. ES: standard ES; CoPES: ours. Post-training methods run for 48 steps. Bold and underlining indicate the best and second-best results, respectively.}
\label{tab:qa_results}
\end{table}

CoPES outperforms standard ES on all nine benchmark--metric combinations, extending its advantage to another agentic task and reward signal. Against full-parameter GRPO, it achieves higher EM and F1 with comparable accuracy on MuSiQue, remains close across metrics on HotpotQA, and trails on 2Wiki. Together with its hardware feasibility, these results support CoPES as a practical option for resource-constrained QA post-training. We additionally evaluate accuracy-based pass@\(k\) on QA; CoPES notably surpasses full-parameter GRPO on 2Wiki and MuSiQue for \(k\geq2\), as detailed in Supplementary~E.

\section{Conclusion}

We introduced Cooperative Parameter-subspace Evolution Strategy~(CoPES), a cooperative coevolutionary method that improves the fixed-budget optimization efficiency of ES-based agentic LLM post-training through parameter-subspace search and joint reward standardization. Experiments on math and multi-hop QA show that CoPES substantially improves over standard ES while approaching full-parameter GRPO, without sacrificing the memory-efficient, full-parameter nature of ES. These findings make ES-based agentic LLM post-training more practical under limited GPU resources. Current experiments cover math on Qwen3.5-4B/9B and multi-hop QA on Qwen3.5-4B; future work will extend CoPES to a broader range of models and agent environments, with particular emphasis on adaptive parameter-space partitioning.

\clearpage

\bibliography{aaai2027}
\clearpage

\setcounter{secnumdepth}{2}
\renewcommand{\thesection}{\Alph{section}}
\renewcommand{\thesubsection}{\thesection.\arabic{subsection}}
\renewcommand{\thefigure}{S\arabic{figure}}
\renewcommand{\thetable}{S\arabic{table}}

\twocolumn[
  \centering
  \textbf{\LARGE Supplementary Material for\\
Cooperative Coevolution for Resource-Constrained Agentic LLM Post-Training}
  \vspace{2em}
]

\section{Additional Experimental Details}
\label{sec:supp_experimental_details}

\subsection{Models and Datasets}
\label{sec:supp_data}

\paragraph{Models.}
We post-train Qwen3.5-4B~\cite{qwen35} on math and multi-hop QA, and Qwen3.5-9B on math. Both models without post-training serve as references; Qwen3.5-9B also provides a larger-model reference for the 4B math comparison.

\paragraph{Math Data.}
We construct the post-training data from the 7,500-example training split of MATH~\cite{hendrycks2021math}. A subject-stratified random split holds out 750 examples for validation. After removing two unusable examples from the remaining data, 6,748 examples are used for post-training. The training and validation data strictly exclude MATH-Test. We evaluate on AIME 2024,\footnote{\url{https://huggingface.co/datasets/HuggingFaceH4/aime_2024}} AIME 2025,\footnote{\url{https://huggingface.co/datasets/test-time-compute/aime_2025}} GSM8K~\cite{cobbe2021gsm8k}, MATH-500~\cite{lightman2024verify}, and MATH-Test. MATH-500 is a curated 500-example subset of the 5,000-example MATH-Test set rather than an independent test set.

\paragraph{QA Data.}
For the QA task, we hold out 750 examples from the HotpotQA training split~\cite{yang2018hotpotqa} for validation and remove them from the post-training data, leaving 89,697 training examples. Evaluation uses 5,000 randomly selected examples from the HotpotQA validation split, 5,000 randomly selected examples from the 2Wiki development split~\cite{ho2020twowiki}, and the complete 2,417-example MuSiQue development split~\cite{trivedi2022musique}. Table~\ref{tab:supp_datasets} summarizes the data used in both tasks.

\begin{table}[t]
\centering
\small
\setlength{\tabcolsep}{1mm}
\begin{tabular}{llllr}
\toprule
Task & Dataset & Source split & Role & \shortstack{Examples\\used} \\
\midrule
\multirow{7}{*}{Math}
& MATH & Training & Post-training & 6,748 \\
& MATH & Training & Validation & 750 \\
& AIME 2024 & Test & Evaluation & 30 \\
& AIME 2025 & Test & Evaluation & 30 \\
& GSM8K & Test & Evaluation & 1,319 \\
& MATH-500 & Test & Evaluation & 500 \\
& MATH-Test & Test & Evaluation & 5,000 \\
\midrule
\multirow{5}{*}{QA}
& HotpotQA & Training & Post-training & 89,697 \\
& HotpotQA & Training & Validation & 750 \\
& HotpotQA & Validation & Evaluation & 5,000 \\
& 2Wiki & Development & Evaluation & 5,000 \\
& MuSiQue & Development & Evaluation & 2,417 \\
\bottomrule
\end{tabular}
\caption{Datasets, source splits, and numbers of examples used for post-training, validation, and evaluation.}
\label{tab:supp_datasets}
\end{table}

\subsection{Agent Environment}
\label{sec:supp_agent}

The two tasks use the same tool-using agent equipped with Local Wiki Search over the Wiki18 corpus~\cite{jin2025searchr1} and a Python Sandbox. The search tool returns evidence from the local corpus, while the sandbox supports commonly used Python packages for computation and answer verification. For each math problem, the agent may invoke Local Wiki Search at most three times and the Python Sandbox at most five times. The corresponding limits for QA are five search calls and five Python calls. The maximum context length is 128K. Invalid tool calls are retried; if a tool or context limit is reached, an available final answer is still extracted and verified, whereas an interaction without a valid answer is treated as unanswered.

\subsection{Training Rewards and Answer Verification}
\label{sec:supp_rewards}

\paragraph{Answer Extraction.}
We inspect assistant outputs in reverse order and first locate the most recent complete answer block, delimited by \texttt{<answer>} and \texttt{</answer>}. Within this block, we use the last boxed or framed value when present; otherwise, we use the entire block content. When no answer block is present, we extract the last boxed value from the full output. If neither form is found, the complete output is used as the fallback prediction. Correctness verification and format scoring are computed separately, so a fallback prediction can still be verified while incurring a format penalty.

\paragraph{Math Reward.}
The extracted math answer is normalized with regular-expression-based cleaning and then checked for mathematical equivalence using SymPy~\cite{meurer2017sympy}. Let \(a_{\mathrm{math}}\in\{0,1\}\) denote the resulting correctness indicator. The training reward is
\begin{equation}
R_{\mathrm{math}}=a_{\mathrm{math}}+r_t+r_f.
\end{equation}
The tool-use bonus is \(r_t=0.1\) when \(a_{\mathrm{math}}=1\) and at least one tool was used, and \(r_t=0\) otherwise.

\paragraph{QA Reward.}
The QA base score is the token-overlap F1 defined in Section~\ref{sec:supp_evaluation}. Its training reward is
\begin{equation}
R_{\mathrm{QA}}=\mathrm{F1}+r_t+r_f,
\end{equation}
where \(r_t=0.1\) if \(\mathrm{F1}>0\) and at least one tool was used, and \(r_t=0\) otherwise.

\paragraph{Format Reward.}
Both tasks use the same format reward. We set \(r_f=0\) when the output contains both an \texttt{<answer>} block and a boxed answer, \(r_f=-0.5\) when exactly one is missing, and \(r_f=-1\) when both are missing.

\subsection{Evaluation Protocol}
\label{sec:supp_evaluation}

\paragraph{Math Evaluation.}
Validation accuracy is computed on the 750 held-out problems using one generation per problem with temperature \(0\). Test-time decoding uses temperature \(0.3\) and top-\(p=1.0\). For GSM8K, MATH-500, and MATH-Test, we generate 32 outputs per problem and report pass@\(k\) for \(k=1,\ldots,16\). For AIME 2024 and AIME 2025, we generate 256 outputs per problem and report \(k=1,\ldots,128\).

\paragraph{QA Evaluation.}
QA validation uses one generation per problem with temperature \(0\), and the validation metric is mean F1 over the 750 held-out problems. Test-time decoding uses temperature \(0.3\), top-\(p=1.0\), and 32 outputs per problem. We normalize predictions and references by lowercasing, removing English articles, stripping and collapsing whitespace, and applying Unicode case and diacritic normalization; punctuation is retained. Accuracy is one when a normalized ground-truth answer is a contiguous substring of the normalized prediction, exact match~(EM) requires equality, and F1 is computed from token overlap. With multiple ground-truth answers, each metric takes its maximum over the references. Each reported Accuracy, EM, or F1 value is first averaged over the 32 outputs for a problem and then averaged across problems.

\paragraph{Pass@\(k\) and Repeated Sampling.}
For a problem with \(n\) generated outputs, of which \(c\) are correct, we use the unbiased estimator~\cite{chen2021codex}
\begin{equation}
\operatorname{pass@}k
=1-\frac{\binom{n-c}{k}}{\binom{n}{k}}.
\label{eq:supp_passk}
\end{equation}
Math correctness is determined by the verifier described above. The additional QA pass@\(k\) results in Section~\ref{sec:supp_qa_passk} use the binary Accuracy criterion. For repeated training runs, these metrics are computed separately for each trained model before aggregation across runs (Section~\ref{sec:stability}).

\subsection{Training Hyperparameters}
\label{sec:supp_hyperparameters}

Table~\ref{tab:supp_hyperparameters} lists the method-specific settings. Following prior work, we use recommended learning-rate settings for the GRPO variants~\cite{shao2024deepseekmath,devulapalli2025clinicalgrpo} and established settings for ES-based LLM post-training~\cite{qiu2026esatscale,liu2025ea4llm}. We determine \(K=4\) before benchmark test evaluation and use the same method-specific hyperparameters on the math and QA tasks. The 9B math experiments retain the 4B optimization hyperparameters. Standard ES and CoPES use greedy generation with temperature \(0\) during perturbed-model evaluation.

\paragraph{Training Budgets and Checkpoint Selection.}
The common math budgets are 16 and 24 updates for Qwen3.5-4B and Qwen3.5-9B, respectively, determined by full-parameter GRPO's best observed validation checkpoints. All three post-training methods in the QA experiment use a common 48-step budget. We validate every eight updates and select the best observed validation checkpoint within the applicable budget. Section~\ref{sec:supp_validation_curves} also reports longer trajectories to characterize training beyond these budgets.

\paragraph{Independent Training Runs.}
The four main 4B math methods are each trained independently three times. The original runs use seed 33 for ES-based methods and seed 42 for GRPO-based methods; the two further runs use seeds 1088 and 2333. Dataset splitting and evaluation-set subsampling use seed 1088 and remain fixed across runs. The main-text tables and pass@\(k\) curves use the original runs; Section~\ref{sec:stability} reports all three runs and their aggregate statistics. The 9B, QA, and ablation results each use one training run per configuration.

\paragraph{Zero Reward Variance.}
ES and CoPES apply no update when the reward standard deviation is at most \(10^{-8}\); the centered rewards are zero when all candidate rewards are equal.

\begin{table}[t]
\centering
\small
\begin{tabular}{ll}
\toprule
Setting & Value \\
\midrule
Prompt batch size & 64 \\
Full-parameter GRPO learning rate & \(1\times10^{-6}\) \\
LoRA-based GRPO learning rate & \(5\times10^{-6}\) \\
GRPO group / mini-batch size & \(8\) / \(32\) \\
GRPO KL coefficient & \(1\times10^{-3}\) \\
GRPO PPO ratio clipping range & \([0.8,1.2]\) \\
GRPO group-sampling temperature & \(1.0\) \\
LoRA rank / scaling factor & \(32\) / \(64\) \\
ES/CoPES population \(N\) & 40 \\
ES/CoPES \(\sigma\) / \(\alpha\) & \(1\times10^{-3}\) / \(5\times10^{-4}\) \\
CoPES \(K\) / \(N_k\) & \(4\) / \(10\) \\
CoPES \(\sigma_k\) & \(2\times10^{-3}\) \\
\bottomrule
\end{tabular}
\caption{Hyperparameters used in the experiments.}
\label{tab:supp_hyperparameters}
\end{table}

\subsection{Frameworks and Compute Environment}
\label{sec:supp_frameworks}

The servers use Intel Xeon Gold 6338 or AMD EPYC 7713 CPUs and run Ubuntu 22.04 with Python 3.11 and CUDA 13. Standard ES and CoPES use vLLM 0.20.2~\cite{kwon2023vllm}, while both GRPO variants use VeRL 0.8.0~\cite{sheng2025hybridflow}.

\subsection{Hardware-Feasibility Configuration}
\label{sec:supp_hardware_feasibility}

For the hardware-feasibility tests reported in the main paper, we enable all applicable memory optimizations supported by each framework. Full-parameter GRPO uses gradient checkpointing together with parameter, optimizer, and activation offloading. LoRA-based GRPO uses gradient checkpointing and parameter and optimizer offloading; activation offloading is unavailable for this configuration. Fused kernels are unavailable for Qwen3.5-4B in the evaluated VeRL configuration. Standard ES and CoPES use the standard vLLM inference path with KV caching, without additional framework-level GPU-memory optimizations.

\subsection{Memory-Efficient ES Implementation}
\label{sec:supp_implementation}

\paragraph{Seed Replay and Chunking.}
Following OpenAI-ES~\cite{salimans2017es}, each perturbation is represented by a random seed and regenerated during the update. CoPES replays each perturbation in parameter chunks, avoiding storage of full perturbation vectors on either GPU or CPU.

\paragraph{CPU Weight Backup.}
Rather than restoring the unperturbed model by subtracting a replayed perturbation, CoPES backs up the pre-perturbation weights in CPU memory and directly restores them after each perturbed-model evaluation. This design avoids residual numerical errors caused by the non-reversibility of floating-point addition and subtraction and ensures that every perturbation is evaluated from the same pre-update model.

\paragraph{Partition Replay.}
CoPES also stores the random seed used to construct each parameter partition. During the update, it replays the chunk-to-subspace assignment from this seed instead of retaining a parameter-level assignment mask. Seed replay, chunked processing, CPU weight backup, and partition replay support memory-efficient execution; they do not alter the algorithmic update described in the main paper.

\subsection{GPU-Hour Profiling}
\label{sec:supp_gpu_hour}

For the 4B experiments, GPU-hour profiling measures training-step time only. We profile CoPES and full-parameter GRPO for 32 steps on the same \(8\times\) RTX 5880 Ada server as representatives of the ES-based and RL-based method classes, respectively, and extrapolate their average per-step costs to each method's reported training steps. A profiled step includes perturbed-model or policy evaluation, tool execution, reward computation, and the parameter update; time spent on validation is excluded. This class-level extrapolation reflects that perturbed-model or policy evaluation dominates per-step time, while partition replay and parameter updates contribute negligibly to total GPU-hours in our measurements. For 9B, all four methods run on the same eight-A100 server and use method-specific log-based timing. Section~\ref{sec:supp_validation_curves} presents both sets of GPU-time curves and the timing procedures.

\begin{table*}[htb]
\centering
\small
\begin{tabular}{lccc}
\toprule
Accounted component & Full-parameter GRPO & LoRA-based GRPO & Standard ES / CoPES \\
\midrule
GPU-resident model and training states & 69.44 GiB & 18.55 GiB & 8.68 GiB \\
Parameter gradients & Included & LoRA parameters only & None \\
Optimizer states & Included & LoRA parameters only & None \\
Backpropagation activations & Included & Included & None \\
Logits and entropy-related tensors & Included & Included & Not retained for backward \\
Context-dependent term & \(3L\) GiB & \(3L\) GiB & \(0.032L\) GiB \\
\midrule
Total & \(69.44+3L\) GiB & \(18.55+3L\) GiB & \(8.68+0.032L\) GiB \\
\bottomrule
\end{tabular}
\caption{Components included in the theoretical GPU memory accounting. \(L\) denotes context length in K tokens.}
\label{tab:supp_memory_accounting}
\end{table*}

\begin{figure}[htbp]
    \centering
    \includegraphics[width=0.75\linewidth]{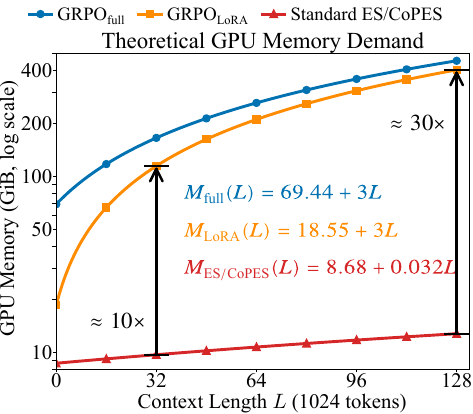}
    \caption{Theoretical GPU memory requirements versus context length with FlashAttention~\cite{dao2022flashattention} and KV caching, excluding other memory optimizations.}
    \label{fig:theoretical_memory}
\end{figure}

\section{Theoretical GPU Memory Accounting}
\label{sec:supp_memory}

This analysis estimates algorithm-level GPU memory requirements for Qwen3.5-4B. It accounts for model weights, parameter gradients, optimizer states, backpropagation activations, logits and entropy-related tensors, and inference KV caches. FlashAttention~\cite{dao2022flashattention} and KV caching are assumed. To isolate the memory implied by each optimization paradigm, the analysis excludes gradient checkpointing, offloading, fused kernels, and other framework-specific memory optimizations. The GRPO estimates cover actor training only and omit a separately resident reference policy and generation engine, making them conservative with respect to the RL-based methods. Standard ES and CoPES have the same theoretical GPU memory requirements.

\paragraph{Fixed States.}
Qwen3.5-4B contains \(P=4{,}659{,}865{,}088\) parameters. Full-parameter GRPO retains FP32 model weights, parameter gradients, and two FP32 AdamW states, giving
\begin{equation}
\frac{(4+4+8)P}{2^{30}}=69.44\ \mathrm{GiB}.
\end{equation}
For LoRA-based GRPO with rank 32, the frozen FP32 base model occupies 17.36 GiB. The \(80{,}150{,}528\) LoRA parameters and their gradients and optimizer states add 1.19 GiB, for a fixed total of 18.55 GiB. ES-based methods retain only the BF16 model weights on GPU:
\begin{equation}
\frac{2P}{2^{30}}=8.68\ \mathrm{GiB}.
\end{equation}
The pre-perturbation backup used by CoPES resides in CPU memory and therefore does not add another GPU-resident model copy.

\paragraph{Context-Dependent Memory.}
For both GRPO variants, the dominant context-dependent terms are the full vocabulary logits, entropy/softmax tensors, and activations retained for backpropagation. Their combined lower-bound estimate is approximately 3 GiB per K tokens. LoRA reduces trainable parameter states but does not remove the need to backpropagate through the model, so it has essentially the same context-dependent slope. ES and CoPES perform forward-only generation and retain neither parameter gradients nor backpropagation activations. Qwen3.5-4B has eight full-attention layers, four KV heads, and a head dimension of 256, yielding 32 KiB of BF16 KV cache per token, or approximately 0.032 GiB per K tokens.

\paragraph{Conservative Rounding.}
The GRPO context-dependent coefficient is rounded near the lower end of its component-wise estimate, whereas the exact ES KV-cache coefficient of 0.03125 GiB per K tokens is rounded upward to 0.032. These choices underestimate rather than exaggerate the relative memory advantage of ES. The values are theoretical accounting estimates rather than measured peak GPU memory and intentionally omit runtime-specific overheads.

\paragraph{Final Estimates.}
The resulting theoretical requirements, with \(L\) measured in units of 1024 tokens, are
\begin{align}
M_{\mathrm{full}}(L) &= 69.44+3L\ \mathrm{GiB}, \\
M_{\mathrm{LoRA}}(L) &= 18.55+3L\ \mathrm{GiB}, \\
M_{\mathrm{ES/CoPES}}(L) &= 8.68+0.032L\ \mathrm{GiB}.
\end{align}
At a context length of 128K, these expressions give 453.44 GiB for full-parameter GRPO, 402.55 GiB for LoRA-based GRPO, and 12.78 GiB for standard ES or CoPES. Thus, even the LoRA-based estimate is more than \(30\times\) the ES estimate, while full-parameter GRPO requires approximately \(35\times\) as much memory. Table~\ref{tab:supp_memory_accounting} summarizes the components, and Figure~\ref{fig:theoretical_memory} shows their scaling with context length.

\begin{figure}[htb]
\centering
\includegraphics[width=0.85\linewidth]{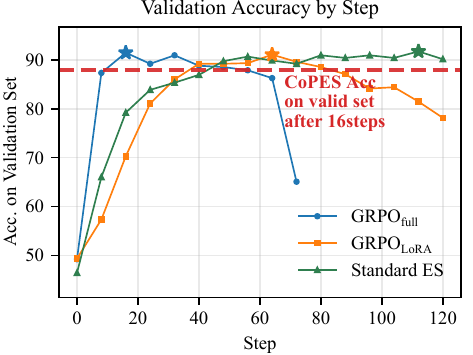}
\caption{Validation accuracy during math post-training. Stars mark the best observed validation checkpoints for full-parameter GRPO, LoRA-based GRPO, and standard ES. The horizontal dashed line marks CoPES validation accuracy at its 16-step fixed-budget endpoint rather than a CoPES training curve.}
\label{fig:supp_math_validation}
\end{figure}

\begin{figure}[htb]
\centering
\includegraphics[width=0.85\linewidth]{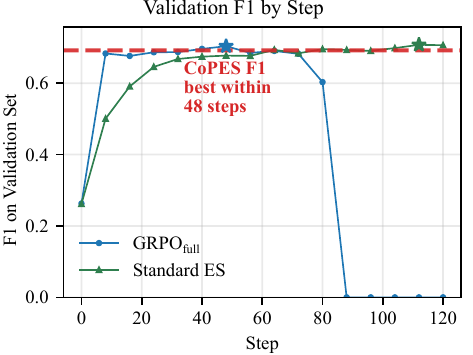}
\caption{Validation F1 during QA post-training. The horizontal dashed line is CoPES validation F1 at its best validation checkpoint within the 48-step budget rather than a CoPES training curve.}
\label{fig:supp_qa_validation}
\end{figure}

\section{Training Dynamics and Compute Efficiency}
\label{sec:supp_validation_curves}

For extended training runs, we report the best observed validation checkpoint. Fixed-budget comparisons select the best observed validation checkpoint within their stated budgets. The extended ES and LoRA trajectories are distinct from the short-budget training runs tabulated in Section~\ref{sec:stability}; their intermediate validation scores can therefore differ. The curves below place these checkpoints in context and relate optimization progress to training compute.

\paragraph{4B Math Validation.}
Figure~\ref{fig:supp_math_validation} shows the Qwen3.5-4B validation trajectories used to identify the best observed validation checkpoints reported in the main paper: step 16 for full-parameter GRPO, step 64 for LoRA-based GRPO, and step 112 for standard ES. The CoPES horizontal dashed line marks the validation accuracy of its original 16-step run. Standard ES first exceeds this reference level at step 48, requiring three times as many training steps as the CoPES fixed-budget run.

\begin{figure}[t]\centering
\includegraphics[width=0.85\linewidth]{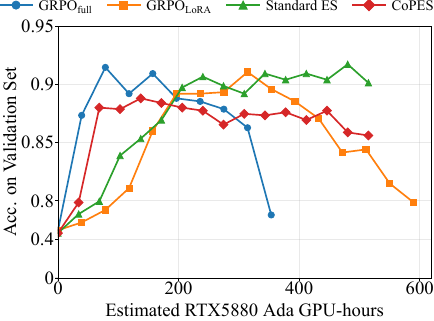}
\caption{Extended 4B validation trajectories versus estimated RTX 5880 Ada GPU-hours. Costs are extrapolated from class-level profiling; the curves use the extended runs rather than the three-run averages.}
\label{fig:gpu_4b}
\end{figure}

\paragraph{Extended CoPES Training.}
CoPES reaches 88.00\% validation accuracy at step 16, close to its 120-step peak of 88.80\% at step 32. Standard ES reaches its peak of 91.73\% at step 112. These trajectories highlight CoPES's rapid early progress under limited training budgets (Figure~\ref{fig:gpu_4b}).

\paragraph{QA Validation.}
Figure~\ref{fig:supp_qa_validation} reports mean validation F1. Full-parameter GRPO reaches its highest observed validation F1 at step 48, which determines the common 48-step QA budget used in the main comparison. The CoPES dashed line marks its validation F1 at the best checkpoint within this budget; its behavior beyond 48 steps was not evaluated. Standard ES reaches the CoPES reference level at step 64; its later checkpoints are shown only to characterize the validation trajectory and are not used in the 48-step main comparison.

\paragraph{Post-Peak Behavior.}
The two GRPO curves on math and the full-parameter GRPO curve on QA decline after reaching their highest observed validation performance. We observed repetitive reasoning loops in some later GRPO outputs, which may contribute to this behavior, but this explanation remains a hypothesis rather than an established causal result.

\paragraph{Estimated GPU-Hours on 4B.}
Figure~\ref{fig:gpu_4b} expresses the extended training trajectories in estimated GPU-hours. The estimates use the same RTX 5880 Ada profiling rates as the main comparison: 4.291 GPU-hours per ES-based update and 4.9132 per GRPO-based update. The figure retains these trajectories' own validation records.

\begin{figure}[t]\centering
\includegraphics[width=0.85\linewidth]{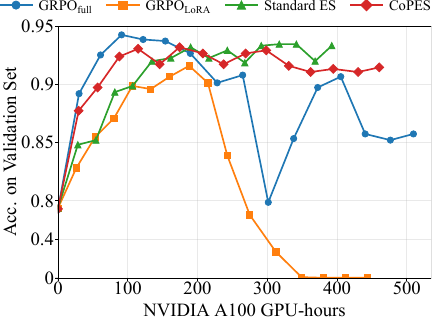}
\caption{Qwen3.5-9B validation accuracy versus A100 GPU-hours on the same eight-GPU server. Costs are obtained from each method\textquotesingle s logs with validation/checkpoint overhead handled as described in the text.}
\label{fig:gpu_9b}
\end{figure}

\paragraph{Per-Method Timing on 9B.}
All four 9B methods run on the same server equipped with eight A100 GPUs. Figure~\ref{fig:gpu_9b} uses method-specific training logs. For ES variants, timestamps delimit training intervals and exclude subsequent validation and checkpoint saving. For GRPO variants, intervals containing validation or saving are replaced by the mean duration of the adjacent training steps to estimate training-only time. Cumulative training time is multiplied by eight GPUs. These curves thus use logged times with the stated overhead correction.

At 24 updates, the resulting training costs are 91.20, 80.16, 81.49, and 88.07 GPU-hours for full-parameter GRPO, LoRA-GRPO, standard ES, and CoPES, respectively. CoPES reaches 92.40\% validation accuracy, compared with 89.33\% for ES and 87.07\% for LoRA-GRPO. The curves compare performance as a function of compute throughout training.

\section{Complete Ablation Results}
\label{sec:supp_ablations}

All ablations use a 16-step fixed budget and a total population of \(N=40\). In the subspace-count ablation, \(N_k=N/K\) and \(\sigma_k=\sqrt{K}\sigma\); the other ablations vary the stated factor while retaining their respective baseline settings. The learning rate is \(\alpha=5\times10^{-4}\), except for ES-\(2\alpha\), which uses \(\alpha=10^{-3}\).

\begin{figure*}[htb]
\centering
\includegraphics[width=0.85\textwidth]{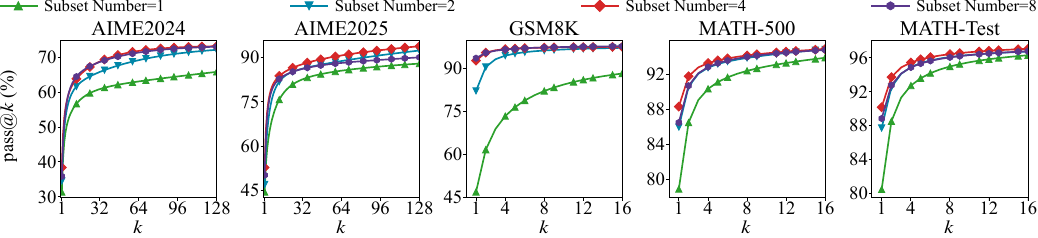}
\caption{Pass@\(k\) for \(K\in\{1,2,4,8\}\), where \(K\) is the number of parameter subspaces. The total population remains \(N=40\), with \(N_k=N/K\) and \(\sigma_k=\sqrt{K}\sigma\).}
\label{fig:supp_subspace_count}
\end{figure*}

\paragraph{Number of Subspaces.}
Figure~\ref{fig:supp_subspace_count} extends the pass@1 ablation in the main paper to the complete pass@\(k\) curves. Increasing \(K\) from 1 to 4 improves pass@1 on all five benchmarks. Increasing it further to 8 does not improve overall performance: \(K=4\) leads on four benchmarks, while \(K=8\) leads on GSM8K. These results support \(K=4\) as the default under this evaluation budget.

\begin{figure*}[htb]
\centering
\includegraphics[width=0.85\textwidth]{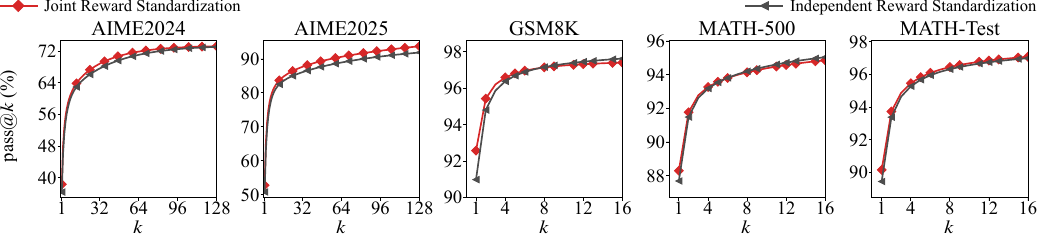}
\caption{Pass@\(k\) with joint reward standardization and independent reward standardization. Both variants use \(K=4\) and otherwise identical settings.}
\label{fig:supp_standardization}
\end{figure*}

\paragraph{Joint Reward Standardization.}
Figure~\ref{fig:supp_standardization} compares the proposed joint standardization with independently standardizing the rewards within each subspace. Joint standardization gives higher pass@1 on all five benchmarks, including gains of 1.94/1.97 percentage points on AIME24/25. At larger \(k\), the GSM8K and MATH-500 curves approach saturation with small crossings. Section~\ref{sec:mechanistic_analysis} analyzes how joint standardization preserves the relative contributions of centered block updates.

\begin{figure*}[!t]
\centering
\includegraphics[width=0.85\textwidth]{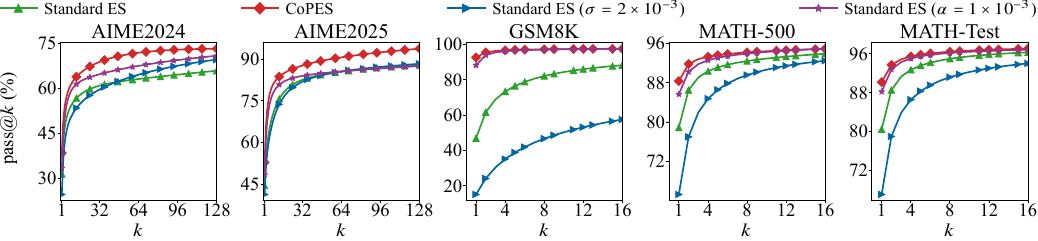}
\caption{Pass@\(k\) for standard ES, CoPES, ES with twice the perturbation scale (ES-\(2\sigma\)), and ES with twice the learning rate (ES-\(2\alpha\)). All methods use the same 16-step budget and total population of 40.}
\label{fig:supp_scale_control}
\end{figure*}

\paragraph{Perturbation-Scale and Learning-Rate Controls.}
CoPES uses \(\sigma_k=2\sigma\) when \(K=4\). Figure~\ref{fig:supp_scale_control} compares CoPES with standard ES and two controls under the same fixed budget: ES-\(2\sigma\) doubles the perturbation scale while retaining the learning rate, and ES-\(2\alpha\) doubles the learning rate while retaining the perturbation scale. Increasing the standard-ES perturbation scale degrades pass@1 across the benchmarks. Doubling its learning rate improves pass@1 over default ES, while CoPES remains ahead on all five benchmarks. Thus, neither adjustment alone reproduces the performance of CoPES.

\section{Additional Results on the QA Task}
\label{sec:supp_qa_passk}

The main paper reports Accuracy, EM, and F1 averaged over the 32 evaluation outputs per problem. Here, we additionally characterize repeated-sampling behavior with Accuracy-based pass@\(k\). A generated output is counted as correct when its binary Accuracy equals one, and Equation~\ref{eq:supp_passk} is applied to the same 32 outputs used to compute the main QA table.

\begin{figure}[htb]
\centering
\includegraphics[width=\linewidth]{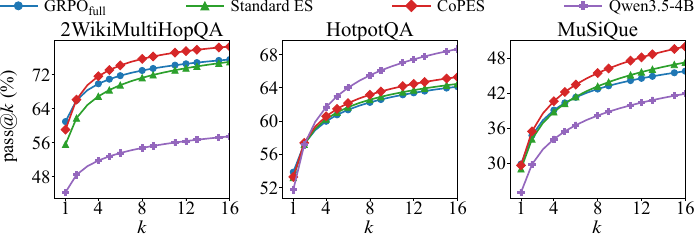}
\caption{Accuracy-based Pass@\(k\) on 2Wiki, HotpotQA, and MuSiQue. Qwen3.5-4B is the reference model without post-training; the other methods use the common 48-step QA budget.}
\label{fig:supp_qa_passk}
\end{figure}

Figure~\ref{fig:supp_qa_passk} reports \(k=1,\ldots,16\). By construction, pass@1 exactly matches the Accuracy values in the main QA table. CoPES exceeds standard ES across the three curves, extending its advantage to repeated sampling on the QA task. CoPES also outperforms full-parameter GRPO at \(k\geq2\) on 2Wiki and MuSiQue. The two methods remain close on HotpotQA, where CoPES also becomes slightly higher as \(k\) increases. Interestingly, Qwen3.5-4B without post-training achieves the highest HotpotQA pass@\(k\) among the evaluated models for \(k=3,\ldots,16\).

\begin{figure*}[t]
\centering
\includegraphics[width=0.85\textwidth]{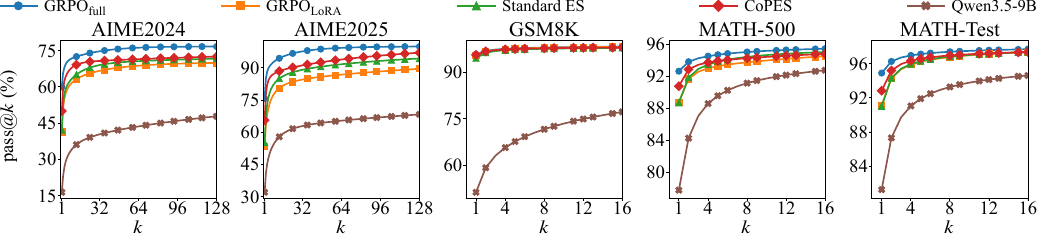}
\caption{Qwen3.5-9B pass@\(k\) under a common 24-step training budget. AIME curves use \(k\le128\); the other benchmarks use \(k\le16\). Increasing \(k\) changes test-time sampling while the trained checkpoints remain fixed.}
\label{fig:passk_9b}
\end{figure*}

\begin{table*}[t]
\centering\small
\setlength{\tabcolsep}{3pt}
\begin{tabular}{llcccccc}\toprule
Method & Run / seed & Valid acc. & AIME24 & AIME25 & GSM8K & MATH-500 & MATH-Test\\\midrule
Full-param GRPO & 42 & 91.47 & 44.28 & 58.68 & 94.19 & 89.71 & 91.66 \\
 & 1088 & 88.93 & 36.69 & 51.88 & 94.37 & 87.81 & 90.24 \\
 & 2333 & 92.67 & 47.38 & 62.16 & 94.33 & 90.26 & 92.45 \\
 & Mean $\pm$ s.d. & $91.02\pm1.91$ & $42.79\pm5.50$ & $57.57\pm5.23$ & $94.30\pm0.10$ & $89.26\pm1.29$ & $91.45\pm1.12$ \\
\midrule
LoRA-GRPO & 42 & 73.60 & 28.14 & 40.76 & 47.49 & 77.19 & 79.18 \\
 & 1088 & 78.00 & 26.65 & 40.82 & 52.09 & 78.57 & 79.98 \\
 & 2333 & 72.80 & 24.61 & 38.41 & 47.90 & 76.85 & 78.52 \\
 & Mean $\pm$ s.d. & $74.80\pm2.80$ & $26.47\pm1.77$ & $40.00\pm1.37$ & $49.16\pm2.55$ & $77.54\pm0.91$ & $79.23\pm0.73$ \\
\midrule
Standard ES & 33 & 77.60 & 31.33 & 44.45 & 46.88 & 78.88 & 80.43 \\
 & 1088 & 76.93 & 31.29 & 44.69 & 63.12 & 81.56 & 83.65 \\
 & 2333 & 84.27 & 31.15 & 44.66 & 78.62 & 85.23 & 87.09 \\
 & Mean $\pm$ s.d. & $79.60\pm4.06$ & $31.25\pm0.10$ & $44.60\pm0.13$ & $62.87\pm15.87$ & $81.89\pm3.19$ & $83.72\pm3.33$ \\
\midrule
CoPES & 33 & 88.00 & 38.39 & 52.75 & 92.57 & 88.29 & 90.16 \\
 & 1088 & 87.47 & 38.72 & 51.21 & 92.90 & 87.66 & 89.99 \\
 & 2333 & 89.87 & 41.03 & 57.01 & 93.43 & 88.27 & 90.48 \\
 & Mean $\pm$ s.d. & $88.44\pm1.26$ & $39.38\pm1.44$ & $53.65\pm3.00$ & $92.97\pm0.43$ & $88.08\pm0.36$ & $90.21\pm0.25$ \\
\bottomrule\end{tabular}
\caption{Three independent Qwen3.5-4B training runs per method under a 16-step budget. For each run, we select the checkpoint with the highest validation accuracy within the budget; all selected checkpoints are at step 16. All scores are percentages. Each method's final row reports the mean and sample standard deviation ($n-1$ denominator) across training runs. Original seeds are 42 for GRPO variants and 33 for ES variants.}
\label{tab:training_runs}
\end{table*}

\section{Additional Results on Qwen3.5-9B}
\label{sec:results_9b}

To examine whether CoPES's gains extend to a larger model, we post-train Qwen3.5-9B using the unchanged 4B optimization hyperparameters. All four methods use a common 24-step budget, set by full-parameter GRPO's best observed validation checkpoint under the same selection rule as the 4B comparison.

Table~4 in the main paper reports pass@1, and Figure~\ref{fig:passk_9b} provides the complete pass@\(k\) curves. CoPES outperforms standard ES and LoRA-GRPO on all five math benchmarks in pass@1, with AIME24/25 gains of 7.84/10.48 percentage points over ES and 8.62/12.34 points over LoRA-GRPO. These results demonstrate that the optimization settings transferred from 4B preserve CoPES's advantage at the larger model scale. Section~\ref{sec:supp_validation_curves} reports the corresponding GPU-time analysis.

\section{Stability Across Independent Training Runs}
\label{sec:stability}
We repeat each of the four Qwen3.5-4B math training procedures with three independent seeds, retaining the same hyperparameters and a 16-step budget. Table~\ref{tab:training_runs} reports every run, followed by its mean and sample standard deviation. Benchmark scores are computed at the validation-selected checkpoint of each run; test results do not determine checkpoint selection.

\paragraph{Consistency of the Advantage.}
CoPES's lowest validation accuracy across runs is 87.47\%, exceeding the highest values for ES (84.27\%) and LoRA-GRPO (78.00\%). The same separation holds on each of the five math benchmarks: the lowest CoPES pass@1 exceeds the highest pass@1 of either baseline. Thus, the advantage persists across all three independently trained models.

\paragraph{Variation across Runs.}
CoPES achieves $88.44\pm1.26\%$ validation accuracy, compared with $79.60\pm4.06\%$ for ES, $74.80\pm2.80\%$ for LoRA-GRPO, and $91.02\pm1.91\%$ for full-parameter GRPO. CoPES has the lowest cross-run validation-accuracy standard deviation among the four methods. The table also reports variability separately for every test benchmark.

\paragraph{Run-Level Statistical Comparison.}
For the directional comparison of CoPES against each of ES and LoRA-GRPO, one-sided exact Wilcoxon rank-sum (Mann--Whitney) tests yield $p=0.05$ for validation accuracy and each math pass@1 metric. Each test uses three independent training runs per method; reported values are unadjusted per-comparison $p$-values. These tests evaluate individual comparisons, while the complete run-level results show the magnitude and consistency of the observed advantage.

\setcounter{equation}{0}
\renewcommand{\theequation}{\thesection.\arabic{equation}}
\section{Mechanistic Analysis of CoPES}
\label{sec:mechanistic_analysis}

CoPES evaluates each block perturbation against a fixed model context and combines the resulting block directions on a common reward scale.
Corollary~1 establishes a direct feedback-attribution property: once a block perturbation is specified, its CoPES candidate has a fixed expected reward, eliminating variation due to random changes outside that block.
Corollary~2 establishes a preservation property of joint standardization: centering retains the block associations in expectation up to a common factor, and shared scaling preserves the relative contributions of the realized centered block vectors.
We connect these properties to the centered ES estimate and the comparison with independent standardization, providing a mechanistic explanation for the gains observed under a fixed evaluation budget.

\subsection{Setup and Directional Reward Association}
\label{sec:mechanistic_setup}

We analyze one training step, conditioning on the current parameters \(\boldsymbol\theta\in\mathbb R^d\), the partition \(\{\mathcal S_k\}_{k=1}^K\), and the shared batch \(\mathcal B\).
Each block has \(d_k=d/K\) coordinates, and \(P_k\in\mathbb R^{d\times d_k}\) embeds a block vector into the full parameter space.
The blocks have disjoint supports, so \(P_k^\top P_k=I_{d_k}\) and \(P_k^\top P_\ell=0\) for \(k\ne\ell\).
With a total of \(N\) candidates, CoPES uses \(N_k=N/K\) candidates per block and the sampling rule
\begin{equation}
\begin{aligned}
\boldsymbol\epsilon_{k,i}&\sim\mathcal N(\boldsymbol0,I_{d_k}),\\
\boldsymbol\theta_{k,i}
&=\boldsymbol\theta+\sigma_kP_k\boldsymbol\epsilon_{k,i},
\qquad \sigma_k=\sqrt K\,\sigma,
\end{aligned}
\label{eq:mechanistic_candidates}
\end{equation}
where \(\sigma\) is the standard-ES perturbation scale.
We assume that \(K\) divides \(d\) and \(N\), with \(N\ge2\).

Let \(r_{k,i}\) denote the observed batch-mean reward and \(J_{\mathcal B}(\boldsymbol\vartheta)\) the expected evaluation reward at parameters \(\boldsymbol\vartheta\).
Thus,
\begin{equation}
\mathbb E[r_{k,i}\mid\boldsymbol\epsilon_{k,i}]
=J_{\mathcal B}(\boldsymbol\theta_{k,i}).
\label{eq:mechanistic_reward_mean}
\end{equation}
Given the conditioning variables above, we assume that different candidate pairs \((\boldsymbol\epsilon_{k,i},r_{k,i})\) are independent, while a candidate's reward can depend on its own perturbation.
Candidates within a block follow the same evaluation protocol; reward distributions may differ across blocks.
The rewards are bounded, so the moments used below exist without assuming that \(J_{\mathcal B}\) is continuous or differentiable.

The reward-weighted update extracts the association between reward and signed perturbation coordinates.
Suppressing the candidate index within block \(k\), define
\begin{equation}
\begin{aligned}
\boldsymbol a_k
&=\mathbb E[(r_k-\mathbb E[r_k])\boldsymbol\epsilon_k]\\
&=\mathbb E[r_k\boldsymbol\epsilon_k]\\
&=\mathbb E_{\boldsymbol\epsilon}
  [J_{\mathcal B}(\boldsymbol\theta+\sigma_kP_k\boldsymbol\epsilon)
   \boldsymbol\epsilon].
\end{aligned}
\label{eq:mechanistic_association}
\end{equation}
The second equality uses \(\mathbb E[\boldsymbol\epsilon_k]=\boldsymbol0\), and the third averages over evaluation randomness first.
Each coordinate of \(\boldsymbol a_k\) is a reward--perturbation covariance, measuring how signed changes in that coordinate are associated with higher or lower reward.\footnote{For example, \(r(\epsilon)=\mathbf1\{|\epsilon|>1\}\) has positive variance under a scalar standard Gaussian but zero \(\mathbb E[r(\epsilon)\epsilon]\), because opposite perturbations have equal expected rewards.}

Gaussian symmetry makes this association interpretable through opposite perturbations.
Changing variables from \(\boldsymbol\epsilon\) to \(-\boldsymbol\epsilon\) in Equation~\eqref{eq:mechanistic_association} and averaging the two expressions gives
\begin{equation}
\begin{aligned}
\boldsymbol a_k=\frac12\mathbb E_{\boldsymbol\epsilon}\Big[
&\big(J_{\mathcal B}(\boldsymbol\theta+\sigma_kP_k\boldsymbol\epsilon)\\
&-J_{\mathcal B}(\boldsymbol\theta-\sigma_kP_k\boldsymbol\epsilon)\big)
\boldsymbol\epsilon\Big].
\end{aligned}
\label{eq:mechanistic_opposite}
\end{equation}
Under independent Gaussian sampling, this identity expresses the association through the expected reward contrast between opposite perturbations.

\subsection{Feedback Attribution with a Fixed Complement}
\label{sec:mechanistic_feedback}

A full-space ES candidate changes both the block being estimated and the parameters used as its evaluation context.
For a standard-ES direction \(\boldsymbol\epsilon_i\sim\mathcal N(\boldsymbol0,I_d)\), write
\begin{equation}
\begin{aligned}
\boldsymbol u_i&=P_k^\top\boldsymbol\epsilon_i,\\
\boldsymbol\delta_{\mathcal S_k^c,i}
&=\sigma(I_d-P_kP_k^\top)\boldsymbol\epsilon_i,
\end{aligned}
\label{eq:mechanistic_complement}
\end{equation}
where \(\mathcal S_k^c\) contains the coordinates outside block \(k\).
Here \(\boldsymbol u_i\in\mathbb R^{d_k}\) is the block direction, whereas \(\boldsymbol\delta_{\mathcal S_k^c,i}\in\mathbb R^d\) is the actual perturbation supported on its complement.
These Gaussian components are independent, and the candidate parameters are \(\boldsymbol\theta+\sigma P_k\boldsymbol u_i+\boldsymbol\delta_{\mathcal S_k^c,i}\).
Let \(r_i\) denote this standard-ES candidate's observed batch-mean reward.

\paragraph{Corollary 1 (Feedback attribution under a fixed complement).}
For any realized partition and any block \(k\), condition on a block direction \(\boldsymbol u\).
The standard-ES candidate can still have variation in its expected reward due to its random complement, whereas the CoPES candidate's expected reward is fully determined:
\begin{equation}
\begin{aligned}
&\operatorname{Var}_{\boldsymbol\delta_{\mathcal S_k^c}}
\!\left[J_{\mathcal B}
(\boldsymbol\theta+\sigma P_k\boldsymbol u
+\boldsymbol\delta_{\mathcal S_k^c})\right]\ge0,\\
&\operatorname{Var}
\!\left[J_{\mathcal B}(\boldsymbol\theta_{k,i})
\mid\boldsymbol\epsilon_{k,i}=\boldsymbol u\right]=0.
\end{aligned}
\label{eq:mechanistic_attribution}
\end{equation}
The first variance is strictly positive whenever the expected reward is not almost surely constant under the complement's sampling distribution.
Thus, fixing the complement eliminates the component of candidate feedback variation caused by random changes outside the selected block.

\textit{Proof.}
The claim follows directly from the candidate constructions.
For standard ES, the complement perturbation remains random after conditioning on \(\boldsymbol u_i=\boldsymbol u\), by independence of the Gaussian components.
Its expected reward therefore has the first variance in Equation~\eqref{eq:mechanistic_attribution}.
For CoPES, conditioning on \(\boldsymbol\epsilon_{k,i}=\boldsymbol u\) fixes the complete candidate to \(\boldsymbol\theta+\sigma_kP_k\boldsymbol u\) by Equation~\eqref{eq:mechanistic_candidates}.
Its expected reward is consequently a constant under this conditioning, proving the zero variance.
\hfill\(\square\)

Corollary~1 makes feedback attribution explicit: each CoPES candidate evaluates the selected block's change within the current full model, with the outside parameters held fixed.
This evaluation includes the block's interactions with the fixed complement.
The result holds conditional on every realized partition, so it applies when the partition is redrawn at each training step.

To connect this attribution property to the update, we next show how complement-induced reward variation enters the centered ES block estimate.
Define
\begin{equation}
\begin{aligned}
m_k(\boldsymbol u)&=\mathbb E[r_i\mid\boldsymbol u_i=\boldsymbol u],\\
\nu_k(\boldsymbol u)&=\operatorname{Var}(r_i\mid\boldsymbol u_i=\boldsymbol u).
\end{aligned}
\label{eq:mechanistic_conditional_reward}
\end{equation}
These quantities average over the complement perturbation and any evaluation randomness.
Let \(U=(\boldsymbol u_1,\ldots,\boldsymbol u_N)\), \(\bar{\boldsymbol u}=N^{-1}\sum_i\boldsymbol u_i\), and \(\mu_r=N^{-1}\sum_i r_i\).
The centered numerator of the ES update in block \(k\) is
\begin{equation}
\boldsymbol h_k^{\mathrm{ES}}
=\frac1N\sum_{i=1}^N(r_i-\mu_r)\boldsymbol u_i.
\label{eq:mechanistic_es_numerator}
\end{equation}
When the reward standard deviation \(s_r>0\), the actual block update is \(\alpha\boldsymbol h_k^{\mathrm{ES}}/s_r\), with the step size \(\alpha>0\) used in the algorithm.

\paragraph{Proposition 1 (Conditional feedback variation).}
Assume that the standard-ES candidate pairs are independent under the same conditioning as above.
Then
\begin{equation}
\mathbb E[\boldsymbol h_k^{\mathrm{ES}}\mid U]
=\frac1N\sum_i m_k(\boldsymbol u_i)
(\boldsymbol u_i-\bar{\boldsymbol u}),
\label{eq:mechanistic_conditional_estimate}
\end{equation}
and
\begin{equation}
\begin{aligned}
&\mathbb E\!\left[
\left\|\boldsymbol h_k^{\mathrm{ES}}
-\mathbb E[\boldsymbol h_k^{\mathrm{ES}}\mid U]\right\|^2
\middle|U\right]\\
&\qquad=\frac1{N^2}\sum_i
\nu_k(\boldsymbol u_i)\|\boldsymbol u_i-\bar{\boldsymbol u}\|^2.
\end{aligned}
\label{eq:mechanistic_conditional_variation}
\end{equation}

\textit{Proof.}
Expanding the sample mean in Equation~\eqref{eq:mechanistic_es_numerator} gives
\[
\boldsymbol h_k^{\mathrm{ES}}
=\frac1N\sum_i r_i(\boldsymbol u_i-\bar{\boldsymbol u}).
\]
Given \(U\), all directions are fixed and candidate independence yields \(\mathbb E[r_i\mid U]=m_k(\boldsymbol u_i)\), proving Equation~\eqref{eq:mechanistic_conditional_estimate}.
Define the remaining reward fluctuation as \(\xi_{k,i}=r_i-m_k(\boldsymbol u_i)\).
Subtracting the conditional mean therefore gives
\[
\boldsymbol h_k^{\mathrm{ES}}
-\mathbb E[\boldsymbol h_k^{\mathrm{ES}}\mid U]
=\frac1N\sum_i\xi_{k,i}(\boldsymbol u_i-\bar{\boldsymbol u}).
\]

The residuals satisfy \(\mathbb E[\xi_{k,i}\mid U]=0\) and \(\mathbb E[\xi_{k,i}^2\mid U]=\nu_k(\boldsymbol u_i)\).
Conditional independence also gives \(\mathbb E[\xi_{k,i}\xi_{k,j}\mid U]=0\) for \(i\ne j\).
Expanding the squared norm and taking its conditional expectation thus leaves only the diagonal terms in Equation~\eqref{eq:mechanistic_conditional_variation}.
\hfill\(\square\)

The candidate reward variance \(\nu_k(\boldsymbol u)\) decomposes into variation from the complement perturbation and evaluation randomness.
Given both \(\boldsymbol u_i\) and \(\boldsymbol\delta_{\mathcal S_k^c,i}\), the complete candidate is fixed and its expected reward is \(J_{\mathcal B}(\boldsymbol\theta+\sigma P_k\boldsymbol u_i+\boldsymbol\delta_{\mathcal S_k^c,i})\).
The law of total variance therefore gives
\begin{equation}
\begin{aligned}
\nu_k(\boldsymbol u)
={}&\operatorname{Var}_{\boldsymbol\delta_{\mathcal S_k^c}}
\!\left[J_{\mathcal B}
(\boldsymbol\theta+\sigma P_k\boldsymbol u
+\boldsymbol\delta_{\mathcal S_k^c})\right]\\
&+\mathbb E_{\boldsymbol\delta_{\mathcal S_k^c}}
\!\left[\operatorname{Var}
(r_i\mid\boldsymbol u_i=\boldsymbol u,
\boldsymbol\delta_{\mathcal S_k^c,i})\right].
\end{aligned}
\label{eq:mechanistic_variance_sources}
\end{equation}
The first term measures changes in expected reward caused by the random complement; the second measures evaluation randomness after all parameters are fixed.
Equation~\eqref{eq:mechanistic_conditional_variation} shows how both sources enter the centered block estimate.

Substituting the first term of Equation~\eqref{eq:mechanistic_variance_sources} into Equation~\eqref{eq:mechanistic_conditional_variation} identifies the nonnegative contribution of random-complement variation to the centered ES block estimate.
Corollary~1 shows that CoPES eliminates this source at candidate evaluation.
Once \(\boldsymbol\epsilon_{k,i}\) is given, its observed reward can still vary through evaluation randomness.
Under deterministic evaluation, that conditional variance is zero.

Fixed-context evaluation can make candidate comparisons more informative about the block being changed when complement-induced fluctuations obscure its response.
In standard ES, complement-induced deviations around the conditional mean \(m_k(\boldsymbol u_i)\) alter reward weights and need not cancel within a finite population.

With a fixed total population \(N\), CoPES allocates \(N/K\) candidates to each \(d/K\)-dimensional block, preserving \(d_k/N_k=d/N\).
Its scale matches the expected squared perturbation norm of standard ES, since \(\sigma_k^2d_k=\sigma^2d\).
This design concentrates each perturbation in one block and evaluates it within the current full-model context.

\subsection{Joint Centering and Shared Scaling}
\label{sec:mechanistic_joint}

Joint standardization gives candidates from different blocks a common reward reference and scale.
We analyze centering in expectation and shared scaling for a fixed candidate population.
For all \(N\) rewards, define
\begin{equation}
\begin{aligned}
\mu_c&=\frac1N\sum_{\ell,j}r_{\ell,j},&
s_c^2&=\frac1N\sum_{\ell,j}(r_{\ell,j}-\mu_c)^2,\\
\boldsymbol h_k&=\frac1{N_k}\sum_i
(r_{k,i}-\mu_c)\boldsymbol\epsilon_{k,i}.
\end{aligned}
\label{eq:mechanistic_joint_statistics}
\end{equation}
The centered numerator \(\boldsymbol h_k\) is defined for every sample outcome, including \(s_c=0\), when all rewards are equal and \(\boldsymbol h_k=\boldsymbol0\).
Expectations of this numerator below include these outcomes; divisions by a standard deviation are used only when it is positive.

The next lemma checks how subtracting the common reward mean affects each block's directional association \(\boldsymbol a_k\).

\paragraph{Lemma 1 (Joint centering).}
For every block,
\begin{equation}
\mathbb E[\boldsymbol h_k]
=\left(1-\frac1N\right)\boldsymbol a_k.
\label{eq:mechanistic_joint_expectation}
\end{equation}
\textit{Proof.}
Let \(b_{k,i}=(N-1)^{-1}\sum_{(\ell,j)\ne(k,i)}r_{\ell,j}\) be the average reward of all other candidates.
Then
\[
r_{k,i}-\mu_c=\frac{N-1}{N}(r_{k,i}-b_{k,i}).
\]
Candidate independence gives \(\mathbb E[b_{k,i}\boldsymbol\epsilon_{k,i}]=\boldsymbol0\), because the baseline is independent of the current direction and that direction has zero mean.
Substituting into the definition of \(\boldsymbol h_k\) and averaging over the \(N_k\) candidates gives
\[
\begin{aligned}
\mathbb E[\boldsymbol h_k]
&=\frac{N-1}{NN_k}\sum_i\boldsymbol a_k\\
&=\left(1-\frac1N\right)\boldsymbol a_k.\quad\square
\end{aligned}
\]

The proof interprets joint centering as comparison with the other candidates' average performance, up to a common factor.
Thus, evaluations in other blocks influence a block's reward weights, while its direction vectors still come from its own \(N_k\) candidates.
The reference is the empirical reward of alternative perturbations, so positive centered weights identify candidates above the population average.

\paragraph{Corollary 2 (Preservation of relative block contributions).}
Under the stated setup, the expected centered numerators \(\mathbb E[\boldsymbol h_k]\) equal the block associations \(\boldsymbol a_k\) multiplied by a common positive factor, preserving their relative magnitudes and all nonzero directions.
For any realized candidate population with \(s_c>0\) and a nonzero full update, shared scaling preserves the direction of each nonzero centered block vector \(\boldsymbol h_k\) and its relative contribution to the full update.

\textit{Proof.}
Lemma~1 proves the first statement because \(1-1/N>0\) is common to all blocks.
For a realized candidate population with \(s_c>0\), let \(\boldsymbol H=\sum_kP_k\boldsymbol h_k\); the algorithm uses\footnote{The shared standard deviation is sample-dependent and remains inside expectations of standardized updates.}
\begin{equation}
\boldsymbol g_k=\frac{\boldsymbol h_k}{s_c},
\qquad
\Delta\boldsymbol\theta=\frac{\alpha}{s_c}\boldsymbol H.
\label{eq:mechanistic_shared_scale}
\end{equation}
Since \(\alpha/s_c>0\), each nonzero block vector retains its direction.
Disjoint supports give \(\|\boldsymbol H\|^2=\sum_k\|\boldsymbol h_k\|^2\), so whenever \(\boldsymbol H\ne\boldsymbol0\),
\begin{equation}
\frac{\|P_k^\top\Delta\boldsymbol\theta\|^2}
     {\|\Delta\boldsymbol\theta\|^2}
=\frac{\|\boldsymbol h_k\|^2}{\sum_\ell\|\boldsymbol h_\ell\|^2}.
\label{eq:mechanistic_relative_contribution}
\end{equation}
This identity proves preservation of each block's relative contribution to the realized update.\hfill\(\square\)

\subsection{Comparison with Independent Standardization}
\label{sec:mechanistic_independent}

To examine the value of the preservation property in Corollary~2, we compare joint and independent standardization using the same candidates.
CoPES evaluates all blocks with the same reward function, batch, block dimension, and perturbation scale, placing their directional responses on a common basis.
The key question is whether separate block standard deviations erase differences in the strength of reward--perturbation associations.

Independent standardization changes both the reward reference and the scale used within each block.
To identify the role of the normalization scale, we first account for the different reward means.
Define the block-specific statistics
\[
\mu_k=\frac1{N_k}\sum_i r_{k,i},
\qquad s_k^2=\frac1{N_k}\sum_i(r_{k,i}-\mu_k)^2,
\]
and
\begin{equation}
\begin{aligned}
\boldsymbol h_k^{\mathrm{ind}}
&=\frac1{N_k}\sum_i(r_{k,i}-\mu_k)\boldsymbol\epsilon_{k,i},\\
\boldsymbol g_k^{\mathrm{ind}}
&=\boldsymbol h_k^{\mathrm{ind}}/s_k\quad(s_k>0).
\end{aligned}
\label{eq:mechanistic_independent_update}
\end{equation}

The two centered numerators differ by an explicit sample-mean term.
Writing \(\bar{\boldsymbol\epsilon}_k=N_k^{-1}\sum_i\boldsymbol\epsilon_{k,i}\), adding and subtracting \(\mu_k\) yields
\begin{equation}
\boldsymbol h_k
=\boldsymbol h_k^{\mathrm{ind}}
+(\mu_k-\mu_c)\bar{\boldsymbol\epsilon}_k.
\label{eq:mechanistic_center_difference}
\end{equation}
For \(s_k>0\), the first term is \(s_k\boldsymbol g_k^{\mathrm{ind}}\).
This decomposition separates the finite-sample effect of the shared reward reference from the effect of the normalization scale.

The expected effects of the two centering rules follow from the same candidate-independence argument.
Expanding \(\mu_k\) gives
\(\mathbb E[\mu_k\boldsymbol\epsilon_{k,i}]=\boldsymbol a_k/N_k\), hence
\begin{equation}
\begin{aligned}
\mathbb E[\boldsymbol h_k^{\mathrm{ind}}]
&=\left(1-\frac1{N_k}\right)\boldsymbol a_k,\\
\mathbb E[(\mu_k-\mu_c)\bar{\boldsymbol\epsilon}_k]
&=\left(\frac1{N_k}-\frac1N\right)\boldsymbol a_k.
\end{aligned}
\label{eq:mechanistic_center_expectations}
\end{equation}
The second identity follows by subtracting the first from Lemma~1; \(\mu_k-\mu_c\) is generally correlated with \(\bar{\boldsymbol\epsilon}_k\).
Because \(N_k=N/K\) is the same for all blocks, the expected numerator under either centering rule is \(\boldsymbol a_k\) multiplied by a coefficient shared across blocks.
We next examine the effect of using a shared standard deviation or separate block standard deviations.

Separate standard deviations make each independently standardized block update invariant to rescaling that block's rewards.
For deterministic \(c_k>0,\beta_k\), consider \(r'_{k,i}=c_kr_{k,i}+\beta_k\) while keeping the directions fixed.
Since \(\mu'_k=c_k\mu_k+\beta_k\) and \(s'_k=c_ks_k\),
\begin{equation}
\frac{r'_{k,i}-\mu'_k}{s'_k}
=\frac{r_{k,i}-\mu_k}{s_k},
\qquad
\boldsymbol g_k^{\mathrm{ind}\prime}
=\boldsymbol g_k^{\mathrm{ind}}.
\label{eq:mechanistic_affine_invariance}
\end{equation}
To track the response amplitude, we now return to expectations of centered numerators.
Under the same reward transformation, the association and jointly centered expectation become
\begin{equation}
\boldsymbol a'_k=c_k\boldsymbol a_k,\qquad
\mathbb E[\boldsymbol h'_k]
=\left(1-\frac1N\right)c_k\boldsymbol a_k.
\label{eq:mechanistic_amplitude}
\end{equation}
Equation~\eqref{eq:mechanistic_amplitude} shows that the expected centered numerators retain these amplitude changes.
Corollary~2 further shows that, for a realized candidate population, division by the shared standard deviation preserves the resulting block vectors' relative contributions to the update.
Joint standardization is invariant to a positive affine transformation shared by all blocks and can respond to relative changes in block reward amplitude.

When block directions and their relative amplitudes carry useful information for the joint update, preserving their relative contributions can improve how that update is allocated across blocks.
For otherwise similar response patterns, independent standardization can rescale a weak response by its own small standard deviation, erasing its magnitude difference from a stronger response.

For a fixed candidate realization, the following example shows how preserving informative response amplitude increases reward at the same actual update length.
Consider two scalar blocks with \(\boldsymbol\theta=(0,0)\), \(N_1=N_2=2\), and \(\sigma_1=\sigma_2=1\), corresponding to \(\sigma=1/\sqrt2\).
Let the deterministic reward be
\[
J_{\mathcal B}(\boldsymbol\vartheta)
=\operatorname{clip}(0.35+0.3\vartheta_1+0.1\vartheta_2,0,1),
\]
where \(\operatorname{clip}(z,0,1)=\min\{1,\max\{0,z\}\}\).
The two parameter coordinates are \(\vartheta_1,\vartheta_2\); before clipping, equal positive displacements produce reward changes in the ratio \(3:1\).
Take the following candidate directions and their rewards, obtained by changing only the indicated coordinate:
\begin{center}
\begin{tabular}{ccc}
\toprule
Block & Directions & Rewards\\
\midrule
1 & \((-0.5,\ 1.5)\) & \((0.2,\ 0.8)\)\\
2 & \((0.5,\ 2.5)\) & \((0.4,\ 0.6)\)\\
\bottomrule
\end{tabular}
\end{center}

These candidates have identical block means, isolating the effect of the standard deviations.
Direct calculation gives
\[
\begin{gathered}
\mu_1=\mu_2=\mu_c=0.5,\\
s_1=0.3,\quad s_2=0.1,\quad s_c=\sqrt{0.05},\\
h_1=h_1^{\mathrm{ind}}=0.3,\qquad
h_2=h_2^{\mathrm{ind}}=0.1.
\end{gathered}
\]
The actual updates with the same \(\alpha\) are
\begin{equation}
\begin{aligned}
\Delta\boldsymbol\theta^{\mathrm{joint}}
&=\frac{\alpha}{\sqrt{0.05}}(0.3,0.1),\\
\Delta\boldsymbol\theta^{\mathrm{ind}}&=\alpha(1,1).
\end{aligned}
\label{eq:mechanistic_example_updates}
\end{equation}
Both have squared length \(2\alpha^2\), without any additional normalization.

For \(0<\alpha\le1\), both updated rewards remain within the unclipped region.
Substituting Equation~\eqref{eq:mechanistic_example_updates} into the reward function gives
\begin{equation}
\begin{aligned}
J_{\mathcal B}(\boldsymbol\theta+\Delta\boldsymbol\theta^{\mathrm{joint}})
-J_{\mathcal B}(\boldsymbol\theta)
&=\frac{\alpha}{\sqrt5},\\
J_{\mathcal B}(\boldsymbol\theta+\Delta\boldsymbol\theta^{\mathrm{ind}})
-J_{\mathcal B}(\boldsymbol\theta)
&=0.4\alpha.
\end{aligned}
\label{eq:mechanistic_example_gains}
\end{equation}
At the same update length, joint standardization produces an approximately \(11.8\%\) larger reward increase in this example, since \((1/\sqrt5)/0.4\approx1.118\).
It preserves the stronger response of the first coordinate, while independent standardization assigns both coordinates the same update magnitude.

The strict reward advantage persists in a sufficiently small neighborhood of the displayed candidates for each fixed \(0<\alpha\le1\).\footnote{Continuity and positive standard deviations preserve the reward ordering. Equality of update lengths holds for the displayed configuration.}

\end{document}